\documentclass[journal]{IEEEtran}
\usepackage{amsmath,amssymb}

\usepackage{multicol}
\usepackage{dcolumn}
\usepackage{graphicx}
\usepackage{subfigure}
\usepackage{float}
\usepackage{epsfig}
\usepackage{amsmath}
\usepackage{amssymb}
\usepackage{array}
\usepackage{epstopdf}
\usepackage{color}
\usepackage{xcolor}
\usepackage[nocompress]{cite}
\usepackage{psfrag}
\usepackage{cite}
\usepackage{multirow}

\graphicspath{{figures/}}
\usepackage[bookmarks,colorlinks,linkcolor=red, anchorcolor=black, citecolor=green]{hyperref}
\usepackage{url}

\begin{document}
%
\title{Beyond a Gaussian Denoiser: Residual Learning of Deep CNN for Image Denoising}
%
%
%


\author{Kai~Zhang,~
        Wangmeng Zuo,
        Yunjin Chen,
        Deyu Meng,
        and~Lei Zhang
\thanks{This project is partially supported by HK RGC GRF grant (under no. PolyU 5313/13E) and the National Natural Scientific Foundation of China (NSFC) under Grant No. 61271093.}
\thanks{K. Zhang is with the School of Computer Science and Technology, Harbin
Institute of Technology, Harbin 150001, China, and also with the Department
of Computing, The Hong Kong Polytechnic University, Hong Kong.}
\thanks{W. Zuo is with the School of Computer Science and
Technology, Harbin Institute of Technology, Harbin 150001, China (e-mail:
cswmzuo@gmail.com).}
\thanks{Y. Chen is with the Institute for Computer Graphics and Vision, Graz
University of Technology, 8010 Graz, Austria.}
\thanks{D. Meng is with the School of Mathematics and Statistics and Ministry of Education Key Lab of Intelligent
Networks and Network Security, Xi'an Jiaotong University, Xi'an 710049, China.}
\thanks{L. Zhang is with the Department of Computing, The Hong Kong
Polytechnic University, Hong Kong (e-mail: cslzhang@comp.polyu.edu.hk).}
}

%
%


\maketitle

\begin{abstract}
Discriminative model learning for image denoising has been recently attracting considerable attentions due to its favorable denoising performance.
In this paper, we take one step forward by investigating the construction of feed-forward denoising convolutional neural networks (DnCNNs) to embrace the progress in very deep architecture, learning algorithm, and regularization method into image denoising. Specifically, residual learning and batch normalization are utilized to speed up the training process as well as boost the denoising performance. Different from the existing discriminative denoising models which usually train a specific model for additive white Gaussian noise (AWGN) at a certain noise level, our DnCNN model is able to handle Gaussian denoising with unknown noise level (i.e., blind Gaussian denoising). With the residual learning strategy, DnCNN implicitly removes the latent clean image in the hidden layers. This property motivates us to train a single DnCNN model to tackle with several general image denoising tasks such as Gaussian denoising, single image super-resolution and JPEG image deblocking. Our extensive experiments demonstrate that our DnCNN model can not only exhibit high effectiveness in several general image denoising tasks, but also be efficiently implemented by benefiting from GPU computing.
\end{abstract}


%
\IEEEpeerreviewmaketitle

\section{Introduction}

Image denoising is a classical yet still active topic in low level vision since it is an indispensable step in many practical applications. The goal of image denoising is to recover a clean image $\mathbf{x}$ from a noisy observation $\mathbf{y}$ which follows an image degradation model $\mathbf{y} = \mathbf{x} + \mathbf{v}$. One common assumption is that $\mathbf{v}$ is additive white Gaussian noise (AWGN) with standard deviation $\sigma$.
From a Bayesian viewpoint, when the likelihood is known, the image prior modeling will play a central role in image denoising. Over the past few decades, various models have been exploited for modeling image priors, including nonlocal self-similarity (NSS) models~\cite{buades2005non,dabov2007image,buades2008nonlocal,mairal2009non}, sparse models~\cite{mairal2009non,elad2006image,dong2013nonlocally}, gradient models~\cite{rudin1992nonlinear,osher2005iterative,weiss2007makes} and Markov random field (MRF) models~\cite{lan2006efficient,li2009markov,roth2009fields}. In particular, the NSS models are popular in state-of-the-art methods such as BM3D~\cite{dabov2007image}, LSSC~\cite{mairal2009non}, NCSR~\cite{dong2013nonlocally} and WNNM~\cite{gu2014weighted}.

Despite their high denoising quality, most of the image prior-based methods typically suffer from two major drawbacks. First, those methods generally involve a complex optimization problem in the testing stage, making the denoising process time-consuming~\cite{dong2013nonlocally,gu2014weighted}. Thus, most of the prior-based methods can hardly achieve high performance without sacrificing computational efficiency. Second, the models in general are non-convex and involve several manually chosen parameters, providing some leeway to boost denoising performance.

To overcome the limitations of prior-based approaches, several discriminative learning methods have been recently
developed to learn image prior models in the context of truncated inference procedure. The resulting models are able to get rid of the iterative optimization procedure in the test phase.
Schmidt and Roth~\cite{schmidt2014shrinkage} proposed a cascade of shrinkage fields (CSF) method that unifies the random field-based model and the unrolled half-quadratic optimization algorithm into a single learning framework. Chen~\emph{et al}.~\cite{chen2015learning,chen2015trainable} proposed a trainable nonlinear reaction diffusion (TNRD) model which learns a modified fields of experts~\cite{roth2009fields} image prior by unfolding a fixed number of gradient descent inference steps. Some of the other related work can be found in~\cite{schmidt2013discriminative,Schmidtpami}.
Although CSF and TNRD have shown promising results toward bridging the gap between computational efficiency and denoising quality, their performance are inherently restricted to the specified forms of prior. To be specific, the priors adopted in CSF and TNRD are based on the analysis model, which is limited in capturing the full characteristics of image structures. In addition, the parameters are learned by stage-wise greedy training plus joint fine-tuning among all stages, and many handcrafted parameters are involved. Another nonnegligible drawback is that they train a specific model for a certain noise level, and are limited in blind image denoising.

In this paper, instead of learning a discriminative model with an explicit image prior, we treat image denoising as a plain discriminative learning problem, i.e., separating the noise from a noisy image by feed-forward convolutional neural networks (CNN).
The reasons of using CNN are three-fold. First, CNN with very deep architecture~\cite{simonyan2014very} is effective in increasing the capacity and flexibility for exploiting image characteristics. Second, considerable advances have been achieved on regularization and learning methods for training CNN, including Rectifier Linear Unit (ReLU)~\cite{krizhevsky2012imagenet}, batch normalization~\cite{ioffe2015batch} and residual learning~\cite{he2015deep}. These methods can be adopted in CNN to speed up the training process and improve the denoising performance. Third, CNN is well-suited for parallel computation on modern powerful GPU, which can be exploited to improve the run time performance.

We refer to the proposed denoising convolutional neural network as DnCNN. Rather than directly outputing the denoised image $\hat{\mathbf{x}}$, the proposed DnCNN is designed to predict the residual image $\hat{\mathbf{v}}$, i.e., the difference between the noisy observation and the latent clean image. In other words, the proposed DnCNN implicitly removes the latent clean image with the operations in the hidden layers. The batch normalization technique is further introduced to stabilize and enhance the training performance of DnCNN. It turns out that residual learning and batch normalization can benefit from each other, and their integration is effective in speeding up the training and boosting the denoising performance.

While this paper aims to design a more effective Gaussian denoiser, we observe that when $\mathbf{v}$ is the difference between the ground truth high resolution image and the bicubic upsampling of the low resolution image, the image degradation model for Guassian denoising can be converted to a single image super-resolution (SISR) problem; analogously, the JPEG image deblocking problem can be modeled by the same image degradation model by taking $\mathbf{v}$ as the difference between the original image and the compressed image. In this sense, SISR and JPEG image deblocking can be treated as two special cases of a ``general'' image denoising problem, though in SISR and JPEG deblocking the noise $\mathbf{vs}$ are much different from AWGN.
It is natural to ask whether is it possible to train a CNN model to handle such general image denoising problem?
By analyzing the connection between DnCNN and TNRD~\cite{chen2015trainable}, we propose to extend DnCNN for handling several general image denoising tasks, including Gaussian denoising, SISR and JPEG image deblocking.

Extensive experiments show that, our DnCNN trained with a certain noise level can yield better Gaussian denoising results than state-of-the-art methods such as BM3D~\cite{dabov2007image}, WNNM~\cite{gu2014weighted} and TNRD~\cite{chen2015trainable}. For Gaussian denoising with unknown noise level (i.e., blind Gaussian denoising), DnCNN with a single model can still outperform BM3D~\cite{dabov2007image} and TNRD~\cite{chen2015trainable} trained for a specific noise level. The DnCNN can also obtain promising results when extended to several general image denoising tasks. Moreover, we show the effectiveness of training only a single DnCNN model for three general image denoising tasks, i.e., blind Gaussian denoising, SISR with multiple upscaling factors, and JPEG deblocking with different quality factors.

The contributions of this work are summarized as follows:
\begin{enumerate}
  \item We propose an end-to-end trainable deep CNN for Gaussian denoising. In contrast to the existing deep neural network-based methods which directly estimate the latent clean image, the network adopts the residual learning strategy to remove the latent clean image from noisy observation.
  \item We find that residual learning and batch normalization can greatly benefit the CNN learning as they can not only speed up the training but also boost the denoising performance. For Gaussian denoising with a certain noise level, DnCNN outperforms state-of-the-art methods in terms of both quantitative metrics and visual quality.
  \item Our DnCNN can be easily extended to handle general image denoising tasks. We can train a single DnCNN model for blind Gaussian denoising, and achieve better performance than the competing methods trained for a specific noise level. Moreover, it is promising to solve three general image denoising tasks, i.e., blind Gaussian denoising, SISR, and JPEG deblocking, with only a single DnCNN model.

\end{enumerate}

The remainder of the paper is organized as follows. Section \ref{sec:related} provides a brief survey of related work. Section \ref{sec:model} first presents the proposed DnCNN model, and then extends it to general image denoising. In Section \ref{sec:experiment}, extensive experiments are conducted to evaluate DnCNNs. Finally, several concluding remarks are given in Section \ref{sec:conclusion}.

\section{Related Work}\label{sec:related}

\subsection{Deep Neural Networks for Image Denoising}

There have been several attempts to handle the denoising problem by deep neural networks. In~\cite{jain2009natural},  Jain and Seung proposed to use convolutional neural networks (CNNs) for image denoising and claimed that CNNs have similar or even better representation power than the MRF model.
In~\cite{burger2012image}, the multi-layer perceptron (MLP) was successfully applied for image denoising.
In~\cite{xie2012image}, stacked sparse denoising auto-encoders method was adopted to handle Gaussian noise removal and achieved comparable results to K-SVD~\cite{elad2006image}.
In~\cite{chen2015trainable}, a trainable nonlinear reaction diffusion (TNRD) model was proposed and it can be expressed as a feed-forward deep network by unfolding a fixed number of gradient descent inference steps.
Among the above deep neural networks based methods, MLP and TNRD can achieve promising performance and are able to compete with BM3D. However, for MLP~\cite{burger2012image} and TNRD~\cite{chen2015trainable}, a specific model is trained for
a certain noise level. To the best of our knowledge, it remains uninvestigated to develop CNN for general image denoising.

\subsection{Residual Learning and Batch Normalization}

Recently, driven by the easy access to large-scale dataset and the advances in deep learning methods, the convolutional neural networks have shown great success in handling various vision tasks. The representative achievements in training CNN models include Rectified Linear Unit (ReLU)~\cite{krizhevsky2012imagenet}, tradeoff between depth and width~\cite{simonyan2014very,szegedy2015googlenet}, parameter initialization~\cite{he2015delving}, gradient-based optimization algorithms~\cite{duchi2011adaptive,zeiler2012adadelta,kingma2014adam}, batch normalization~\cite{ioffe2015batch} and residual learning~\cite{he2015deep}. Other factors, such as the efficient training implementation on modern powerful GPUs, also contribute to the success of CNN. For Gaussian denoising, it is easy to generate sufficient training data from a set of high quality images. This work focuses on the design and learning of CNN for image denoising. In the following, we briefly review two methods related to our DnCNN, i.e., residual learning and batch normalization.

\subsubsection{Residual Learning}

Residual learning~\cite{he2015deep} of CNN was originally proposed to solve the performance degradation problem, i.e., even the training accuracy begins to degrade along with the increasing of network depth. By assuming that the residual mapping is much easier to be learned than the original unreferenced mapping, residual network explicitly  learns a residual mapping for a few stacked layers. With such a residual learning strategy, extremely deep CNN can be easily trained and improved accuracy has been achieved for image classification and object detection~\cite{he2015deep}.

The proposed DnCNN model also adopts the residual learning formulation. Unlike the residual network~\cite{he2015deep} that uses many residual units (i.e., identity shortcuts), our DnCNN employs a single residual unit to predict the residual image. We further explain the rationale of residual learning formulation by analyzing its connection with TNRD~\cite{chen2015trainable}, and extend it to solve several general image denoising tasks. It should be noted that, prior to the residual network~\cite{he2015deep}, the strategy of predicting the residual image has already been adopted in some low-level vision problems such as single image super-resolution~\cite{timofte2014a} and color image demosaicking~\cite{kiku2013residual}.
However, to the best of our knowledge, there is no work which directly predicts the residual image for denoising.

\subsubsection{Batch Normalization}

Mini-batch stochastic gradient descent (SGD) has been widely used in training CNN models. Despite the simplicity and effectiveness of mini-batch SGD, its training efficiency is largely reduced by internal covariate shift~\cite{ioffe2015batch}, i.e., changes in the distributions of internal nonlinearity inputs during training. Batch normalization~\cite{ioffe2015batch} is proposed to alleviate the internal covariate shift by incorporating a normalization step and a scale and shift step before the nonlinearity in each layer. For batch normalization, only two parameters per activation are added, and they can be updated with back-propagation.  Batch normalization enjoys several merits, such as fast training, better performance, and low sensitivity to initialization. For further details on batch normalization, please refer to~\cite{ioffe2015batch}.

By far, no work has been done on studying batch normalization for CNN-based image denoising. We empirically find that,
the integration of residual learning and batch normalization can result in fast and stable training and better denoising performance.

\section{The Proposed Denoising CNN Model} \label{sec:model}

\begin{table*}[htbp]
\caption{The effective patch sizes of different methods with noise level $\sigma = 25$.}
\center
\begin{tabular}{|c|c|c|c|c|c|c|}
  \hline
  Methods & BM3D~\cite{dabov2007image}&  WNNM~\cite{gu2014weighted}& EPLL~\cite{zoran2011learning}& MLP~\cite{burger2012image}  & CSF~\cite{schmidt2014shrinkage} & TNRD~\cite{chen2015trainable} \\ \hline
  Effective Patch Size & $49 \times 49$ &  $361 \times 361$&$36 \times 36$ & $47 \times 47$ & $61 \times 61$ & $61 \times 61$   \\
  \hline
\end{tabular}
\label{table11}
\end{table*}

\begin{figure*}[htbp]
  \centering
  \includegraphics[width=0.98\textwidth]{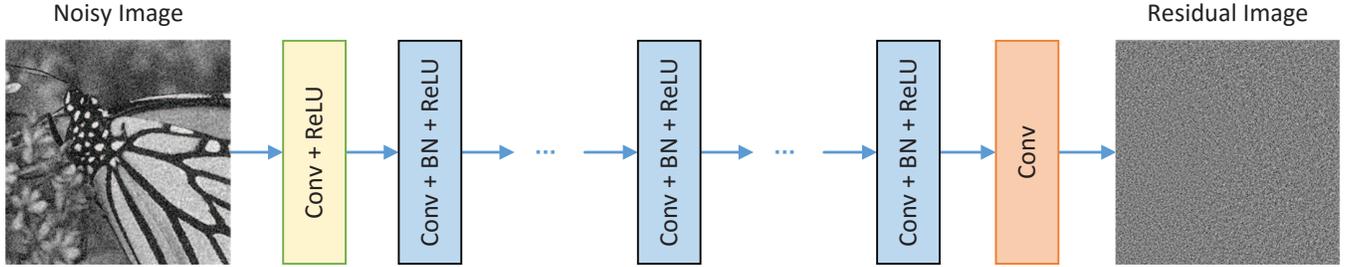}\\
  \caption{The architecture of the proposed DnCNN network.}\label{fig1}\vspace{-0.3cm}
\end{figure*}

In this section, we present the proposed denoising CNN model, i.e., DnCNN, and extend it for handling several general image denoising tasks. Generally, training a deep CNN model for a specific task generally involves two steps: ($i$) network architecture design and ($ii$) model learning from training data. For network architecture design, we modify the VGG network~\cite{simonyan2014very} to make it suitable for image denoising, and set the depth of the network based on the effective patch sizes used in state-of-the-art denoising methods. For model learning, we adopt the residual learning formulation, and incorporate it with batch normalization for fast training and improved denoising performance. Finally, we discuss the connection between DnCNN and TNRD~\cite{chen2015trainable}, and extend DnCNN for {several general image denoising tasks}.

\subsection{Network Depth}

Following the principle in~\cite{simonyan2014very}, we set the size of convolutional filters to be $3 \times 3$ but remove all pooling layers. Therefore, the receptive field of DnCNN with depth of $d$ should be $(2d+1)\times(2d+1)$. Increasing receptive field size can make use of the context information in larger image region. For better tradeoff between performance and efficiency, one important issue in architecture design is to set a proper depth for DnCNN.

It has been pointed out that the receptive field size of denoising neural networks correlates with the effective patch size of denoising methods~\cite{jain2009natural,burger2012image}. Moreover, high noise level usually requires larger effective patch size to capture more context information for restoration~\cite{levin2011natural}. Thus, by fixing the noise level $\sigma = 25$, we analyze the effective patch size of several leading denoising methods to guide the depth design of our DnCNN. In BM3D~\cite{dabov2007image}, the non-local similar patches are adaptively searched in a local widow of size $25 \times 25$ for two times, and thus the final effective patch size is $49 \times 49$. Similar to BM3D, WNNM~\cite{gu2014weighted} uses a larger searching window and performs non-local searching iteratively, resulting in a quite large effective patch size ($361 \times 361$). MLP~\cite{burger2012image} first uses a patch of size $39 \times 39$ to generate the predicted patch, and then adopts a filter of size $9 \times 9$ to average the output patches, thus its effective patch size is $47 \times 47$. The CSF~\cite{schmidt2014shrinkage} and TNRD~\cite{chen2015trainable} with five stages involves a total of ten convolutional layers with filter size of $7 \times 7$, and their effective patch size is $61 \times 61$.

Table~\ref{table11} summarizes the effective patch sizes adopted in different methods with noise level $\sigma = 25$.
It can be seen that the effective patch size used in EPLL~\cite{zoran2011learning} is the smallest, i.e., $36 \times 36$. It is interesting to verify whether DnCNN with the receptive field size similar to EPLL can compete against the leading denoising methods. Thus, for Gaussian denoising with a certain noise level, we set the receptive field size of DnCNN to $35 \times 35$ with the corresponding depth of 17. For other general image denoising tasks, we adopt a larger receptive field and set the depth to be 20.

\subsection{Network Architecture}

The input of our DnCNN is a noisy observation $\mathbf{y} = \mathbf{x} + \mathbf{v}$. Discriminative denoising models such as MLP~\cite{burger2012image} and CSF~\cite{schmidt2014shrinkage} aim to learn a mapping function $\mathcal{F}(\mathbf{y}) = \mathbf{x}$ to predict the latent clean image. For DnCNN, we adopt the residual learning formulation to train a residual mapping $\mathcal{R}(\mathbf{y}) \approx \mathbf{v}$, and then we have $\mathbf{x} = \mathbf{y} - \mathcal{R}(\mathbf{y})$.
Formally, the averaged mean squared error between the desired residual images and estimated ones from noisy input
\begin{equation}\label{eq:loss}
  \ell(\mathbf{\Theta}) = \frac{1}{2N}\sum_{i=1}^N\|\mathcal{R}(\mathbf{y}_i; \mathbf{\Theta}) - (\mathbf{y}_i - \mathbf{x}_i) \|_F^2
\end{equation}
can be adopted as the loss function to learn the trainable parameters $\mathbf{\Theta}$ in DnCNN. Here $\{(\mathbf{y}_i, \mathbf{x}_i)\}_{i=1}^N$ represents $N$ noisy-clean training image (patch) pairs. Fig.~\ref{fig1} illustrates the architecture of the proposed DnCNN for learning $\mathcal{R}(\mathbf{y})$. In the following, we explain the architecture of DnCNN and the strategy for reducing boundary artifacts.

\subsubsection{Deep Architecture}
Given the DnCNN with depth $D$, there are three types of layers, shown in Fig.~\ref{fig1} with three different colors. ($i$) Conv+ReLU: for the first layer, 64 filters of size 
$3 \times 3 \times c$ are used to generate 64 feature maps, and rectified linear units (ReLU, $max(0, \cdot)$) are then utilized for nonlinearity. Here $c$ represents the number of image channels, i.e., $c=1$ for gray image and $c=3$ for color image. ($ii$) Conv+BN+ReLU: for layers 2 $\sim$ $(D-1)$, 64 filters of size $3 \times 3 \times 64$ are used, and batch normalization~\cite{ioffe2015batch} is added  between convolution and ReLU. ($iii$) Conv: for the last layer, $c$ filters of size $3 \times 3 \times 64$ are used to reconstruct the output.

To sum up, our DnCNN model has two main features: the residual learning formulation is adopted to learn $\mathcal{R}(\mathbf{y})$, and batch normalization is incorporated to speed up training as well as boost the denoising performance.
By incorporating convolution with ReLU, DnCNN can gradually separate image structure from the noisy observation through the hidden layers. Such a mechanism is similar to the iterative noise removal strategy adopted in methods such as EPLL and WNNM, but our DnCNN is trained in an end-to-end fashion.
Later we will give more discussions on the rationale of combining residual learning and batch normalization.

\subsubsection{Reducing Boundary Artifacts}
In many low level vision  applications, it usually requires that the output image size should keep the same as the input one. This may lead to the boundary artifacts. In MLP~\cite{burger2012image}, boundary of the noisy input image is symmetrically padded in the preprocessing stage, whereas the same padding strategy is carried out before every stage in CSF~\cite{schmidt2014shrinkage} and TNRD~\cite{chen2015trainable}. Different from the above methods, we directly pad zeros before convolution to make sure that each feature map of the middle layers has the same size as the input image. We find that the simple zero padding strategy does not result in any boundary artifacts. This good property is probably attributed to the powerful ability of the DnCNN.

\subsection{Integration of Residual Learning and Batch Normalization for Image Denoising}

\begin{figure*}[!htbp]
  \centering
\subfigure[SGD]
{\includegraphics[width=0.48\textwidth]{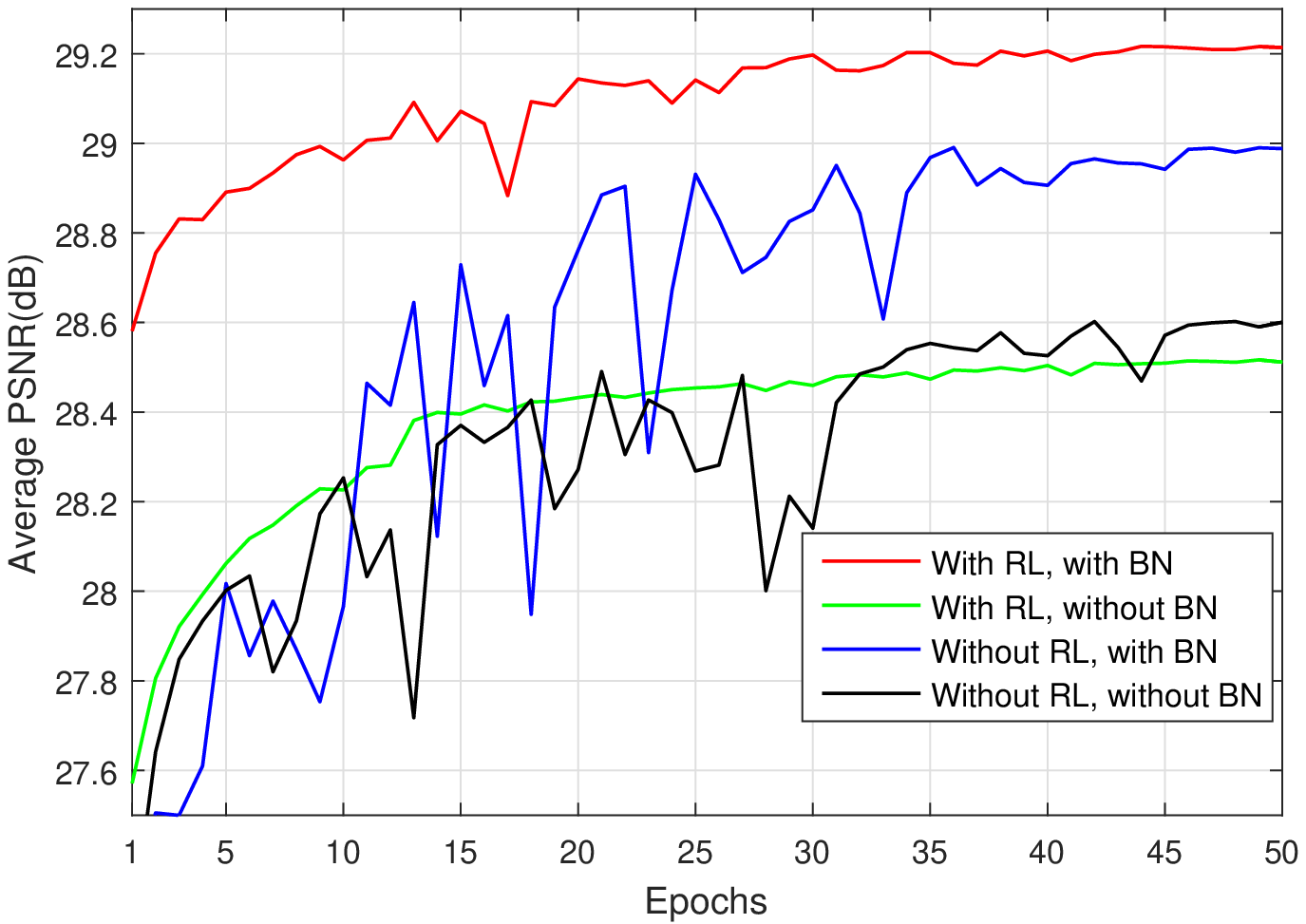}}
\subfigure[Adam]
{\includegraphics[width=0.48\textwidth]{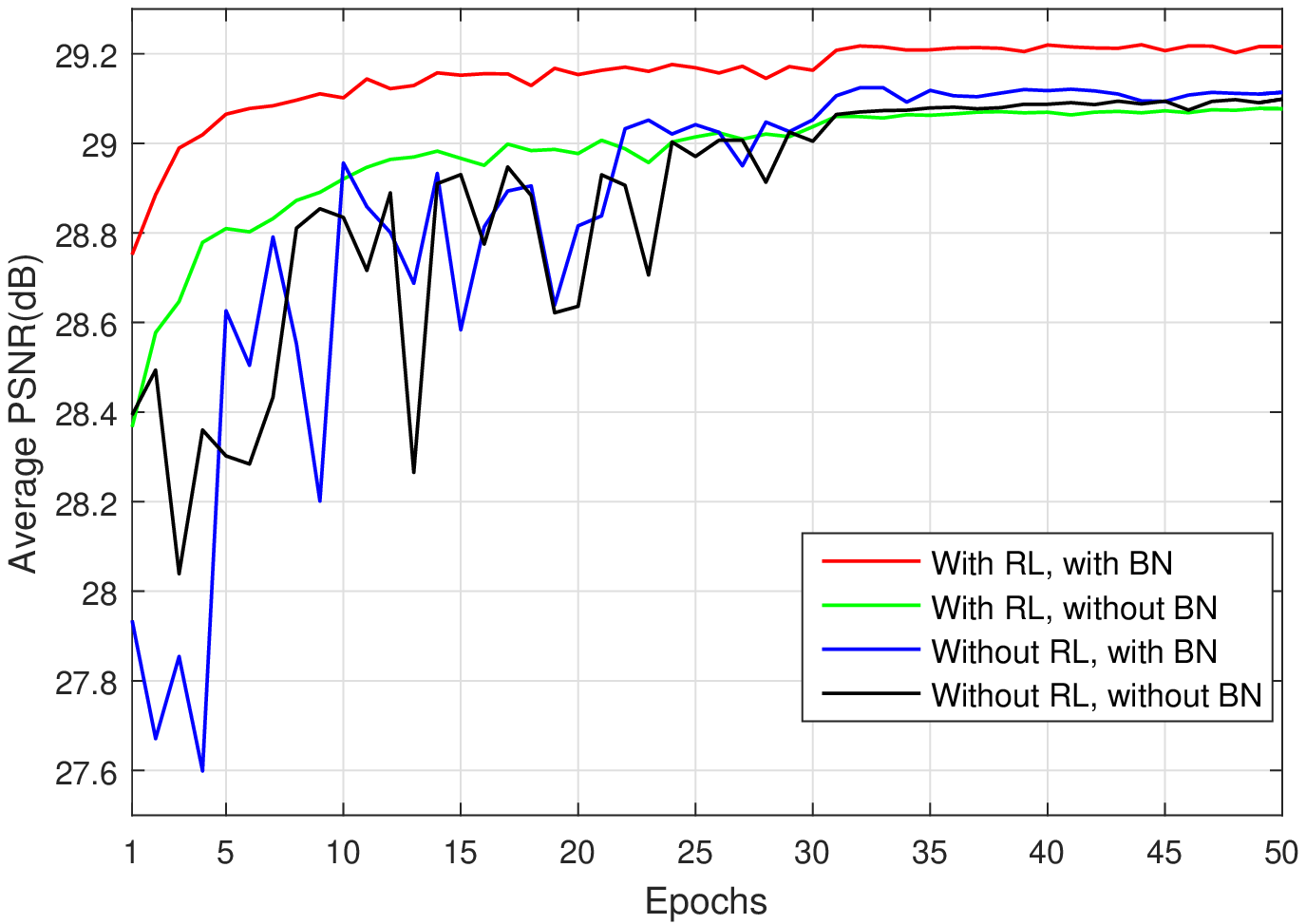}}
  \caption{The Gaussian denoising results of four specific models under two gradient-based optimization algorithms, i.e., (a) SGD, (b) Adam, with respect to epochs. The four specific models are in different combinations of residual learning (RL) and batch normalization (BN) and are trained with noise level 25. The results are evaluated on 68 natural images from Berkeley segmentation dataset.}\label{fig_sgd_adam}
\end{figure*}

The network shown in Fig.~\ref{fig1} can be used to train either the original mapping $\mathcal{F}(\mathbf{y})$ to predict $\mathbf{x}$ or the residual mapping $\mathcal{R}(\mathbf{y})$ to predict $\mathbf{v}$. According to~\cite{he2015deep}, when the original mapping is more like an identity mapping, the residual mapping will be much easier to be optimized. Note that the noisy observation $\mathbf{y}$ is much more like the latent clean image $\mathbf{x}$ than the residual image $\mathbf{v}$ (especially when the noise level is low). Thus, $\mathcal{F}(\mathbf{y})$ would be more close to an identity mapping than $\mathcal{R}(\mathbf{y})$, and the residual learning formulation is more suitable for image denoising.

Fig.~\ref{fig_sgd_adam} shows the average PSNR values obtained using these two learning formulations with/without batch normalization under the same setting on gradient-based optimization algorithms and network architecture. Note that two gradient-based optimization algorithms are adopted: one is the stochastic gradient descent algorithm with momentum (i.e., SGD) and the other one is the Adam algorithm~\cite{kingma2014adam}.
Firstly, we can observe that the residual learning formulation can result in faster and more stable convergence than the original mapping learning. In the meanwhile, without batch normalization, simple residual learning with conventional SGD cannot compete with the state-of-the-art denoising methods such as TNRD (28.92dB).
We consider that the insufficient performance should be attributed to the internal covariate shift~\cite{ioffe2015batch} caused by the changes in network parameters during training.
Accordingly, batch normalization is adopted to address it.
Secondly, we observe that, with batch normalization, learning residual mapping (the red line) converges faster and exhibits better denoising performance than learning original mapping (the blue line). In particular, both the SGD and Adam optimization algorithms can enable the network with residual learning and batch normalization to have the best results.
In other words, it is the integration of residual learning formulation and batch normalization rather than the optimization algorithms (SGD or Adam) that leads to the best denoising performance.

Actually, one can notice that in Gaussian denoising the residual image and batch normalization are both associated with the Gaussian distribution. It is very likely that residual learning and batch normalization can benefit from each other for Gaussian denoising\footnote{It should be pointed out that this does not mean that our DnCNN can not handle other general denoising tasks well.}.
This point can be further validated by the following analyses.

\begin{itemize}
  \item On the one hand, residual learning benefits from batch normalization. This is straightforward because batch normalization offers some merits for CNNs, such as alleviating internal covariate shift problem. From Fig.~\ref{fig_sgd_adam}, one can see that even though residual learning without batch normalization (the green line) has a fast convergence, it is inferior to residual learning with batch normalization (the red line).
  \item On the other hand, batch normalization benefits from residual learning. As shown in Fig.~\ref{fig_sgd_adam}, without residual learning, batch normalization even has certain adverse effect to the convergence (the blue line). With residual learning, batch normalization can be utilized to speedup the training as well as boost the performance (the red line). Note that each mini-bath is a small set (e.g., 128) of images. Without residual learning, the input intensity and the convolutional feature are correlated with their neighbored ones, and the distribution of the layer inputs also rely on the content of the images in each training mini-batch. With residual learning, DnCNN implicitly removes the latent clean image with the operations in the hidden layers. This makes that the inputs of each layer are Gaussian-like distributed, less correlated, and less related with image content. Thus, residual learning can also help batch normalization in reducing internal covariate shift.

\end{itemize}

To sum up, the integration of residual learning and batch normalization can not only speed up and stabilize the training process but also boost the denoising performance.

\subsection{Connection with TNRD} \label{sec:tnrd}
Our DnCNN can also be explained as the generalization of one-stage TNRD~\cite{chen2015learning,chen2015trainable}. Typically, TNRD aims to train a discriminative solution for the following problem
\begin{eqnarray} \label{eq:tnrd}
\min_{ \mathbf{x} }  \Psi( \mathbf{y} - \mathbf{x} ) + \lambda\sum_{k=1}^{K} \sum_{p=1}^{N} \rho_k ( (\mathbf{f}_k \ast \mathbf{x} )_p ),
\end{eqnarray}
from an abundant set of degraded-clean training  image pairs. Here $N$ denotes the image size, $\lambda$ is the regularization parameter, $\mathbf{f}_k \ast \mathbf{x}$ stands for the convolution of the image $\mathbf{x}$ with the $k$-th filter kernel $\mathbf{f}_k$, and $\rho_k(\cdot)$ represents the $k$-th penalty function which is adjustable in the TNRD model. For Gaussian denoising, we set $\Psi( \mathbf{z}  ) = \frac{1}{2} \| \mathbf{z} \| ^{2}$.

The diffusion iteration of the first stage can be interpreted as performing one gradient descent inference step at starting point $\mathbf{y}$, which is given by
\begin{eqnarray} \label{eq:diffusion1}
\mathbf{x}_1 =  \mathbf{y} - \alpha\lambda\sum_{k=1}^{K} (\bar{\mathbf{f}}_k \ast \phi_k (\mathbf{f}_k \ast  \mathbf{y}  ) ) - \alpha \left. \frac{\partial \Psi( \mathbf{z} )}{\partial \mathbf{z} } \right|_{\mathbf{z} = \mathbf{0}} ,
\end{eqnarray}
where $\bar{\mathbf{f}}_k$ is the adjoint filter of $\mathbf{f}_k$ (i.e., $\bar{\mathbf{f}}_k$ is obtained
by rotating 180 degrees the filter $\mathbf{f}_k$), $\alpha$ corresponds to the stepsize and $\rho^{\prime}_{k}(\cdot)= \phi_k(\cdot)$.
For Gaussian denoising, we have $\left. \frac{\partial \Psi( \mathbf{z} )}{\partial \mathbf{z} } \right|_{\mathbf{z} = \mathbf{0}} = \mathbf{0}$, and Eqn.~\eqref{eq:diffusion1} is equivalent to the following expression
\begin{eqnarray} \label{eq:diffusion2}
\mathbf{v}_1 = \mathbf{y} - \mathbf{x}_1 = \alpha\lambda\sum_{k=1}^{K} (\bar{\mathbf{f}}_k \ast \phi_k (\mathbf{f}_k \ast  \mathbf{y}  ) ) ,
\end{eqnarray}
where $\mathbf{v}_1$ is the estimated residual of $\mathbf{x}$ with respect to $\mathbf{y}$.

Since the influence function $\phi_k(\cdot)$ can be regarded as point-wise nonlinearity applied to convolution feature maps, Eqn.~\eqref{eq:diffusion2} actually is a two-layer feed-forward CNN.
As can be seen from Fig.~\ref{fig1}, the proposed CNN architecture further generalizes one-stage TNRD from three aspects: ($i$) replacing the influence function with ReLU to ease CNN training; ($ii$) increasing the CNN depth to improve the capacity in modeling image characteristics; ($iii$) incorporating with batch normalization to boost the performance. The connection with one-stage TNRD provides insights in explaining the use of residual learning for CNN-based image restoration. Most of the parameters in Eqn.~\eqref{eq:diffusion2} are derived from the analysis prior term of Eqn.~\eqref{eq:tnrd}. In this sense, most of the parameters in DnCNN are representing the image priors.

It is interesting to point out that, even the noise is not Gaussian distributed (or the noise level of Gaussian is unknown), we still can utilize Eqn.~\eqref{eq:diffusion1} to obtain $\mathbf{v}_1$ if we have
\begin{eqnarray} \label{eq:Psi}
\left. \frac{\partial \Psi( \mathbf{z} )}{\partial \mathbf{z} } \right|_{\mathbf{z} = \mathbf{0}} = \mathbf{0}.
\end{eqnarray}  Note that Eqn.~\eqref{eq:Psi} holds for many types of noise distributions, e.g., generalized Gaussian distribution. It is natural to assume that it also holds for the noise caused by SISR and JPEG compression. It is possible to train a single CNN model for {several general image denoising tasks}, such as Gaussian denoising with unknown noise level, SISR with multiple upscaling factors, and JPEG deblocking with different quality factors.

Besides, Eqn.~\eqref{eq:diffusion2} can also be interpreted as the operations to remove the latent clean image $\mathbf{x}$ from the degraded observation $\mathbf{y}$ to estimate the residual image $\mathbf{v}$. For {these tasks}, even the noise distribution is complex, it can be expected that our DnCNN would also perform robustly in predicting residual image by gradually removing the latent clean image in the hidden layers.

\subsection{Extension to General Image Denoising}

The existing discriminative Gaussian denoising methods, such as MLP, CSF and TNRD, all train a specific model for a fixed noise level~\cite{chen2015trainable,burger2012image}. When applied to Gaussian denoising with unknown noise, one common way is to first estimate the noise level, and then use the model trained with the corresponding noise level. This makes the denoising results affected by the accuracy of noise estimation. In addition, those methods cannot be applied to the cases with {non-Gaussian noise distribution}, e.g., SISR and JPEG deblocking.

Our analyses in Section \ref{sec:tnrd} have shown the potential of DnCNN in general image denoising. To demonstrate it, we first extend our DnCNN for Gaussian denoising with unknown noise level. In the training stage, we use the noisy images from a wide range of noise levels (e.g., $\sigma \in [0, 55]$) to train a single DnCNN model. Given a test image whose noise level belongs to the noise level range, the learned single DnCNN model can be utilized to denoise it without estimating its noise level.

We further extend our DnCNN by learning a single model for {several general image denoising tasks}. We consider three specific tasks, i.e., blind Gaussian denoising, SISR, and JPEG deblocking. In the training stage, we utilize the images with AWGN from a wide range of noise levels, down-sampled images with multiple upscaling factors, and JPEG images with different quality factors to train a single DnCNN model. Experimental results show that the learned single DnCNN model is able to yield excellent results for any of the {three general image denoising tasks}.

\section{Experimental Results}\label{sec:experiment}
\label{sec:blind}

\subsection{Experimental setting}

\subsubsection{Training and Testing Data}

For Gaussian denoising with either known or unknown noise level, we follow~\cite{chen2015trainable} to use 400 images of size $180\times180$ for training. We found that using a larger training dataset can only bring negligible improvements. To train DnCNN for Gaussian denoising with known noise level, we consider three noise levels, i.e., $\sigma=15$, $25$ and $50$.
We set the patch size as $40\times40$, and crop $128 \times 1,600$ patches to train the model. We refer to our DnCNN model for Gaussian denoising with known specific noise level as DnCNN-S.

To train a single DnCNN model for blind Gaussian denoising, we set the range of the noise levels as $\sigma \in [0, 55]$, and the patch size as $50\times50$. $128\times3,000$ patches are cropped to train the model. We refer to our single DnCNN model for blind Gaussian denoising task as DnCNN-B.

\begin{figure*}[!htbp]
\begin{center}
\subfigure
{\includegraphics[width=0.077\textwidth]{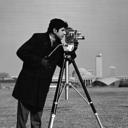}}
\subfigure
{\includegraphics[width=0.077\textwidth]{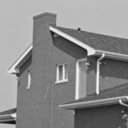}}
\subfigure
{\includegraphics[width=0.077\textwidth]{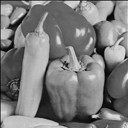}}
\subfigure
{\includegraphics[width=0.077\textwidth]{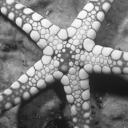}}
\subfigure
{\includegraphics[width=0.077\textwidth]{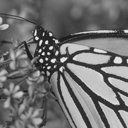}}
\subfigure
{\includegraphics[width=0.077\textwidth]{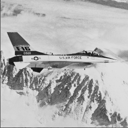}}
\subfigure
{\includegraphics[width=0.077\textwidth]{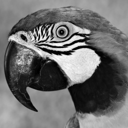}}
\subfigure
{\includegraphics[width=0.077\textwidth]{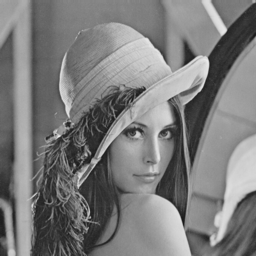}}
\subfigure
{\includegraphics[width=0.077\textwidth]{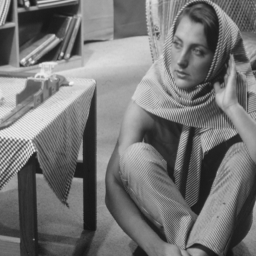}}
\subfigure
{\includegraphics[width=0.077\textwidth]{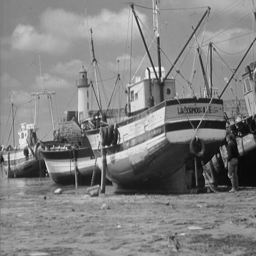}}
\subfigure
{\includegraphics[width=0.077\textwidth]{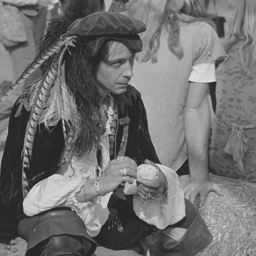}}
\subfigure
{\includegraphics[width=0.077\textwidth]{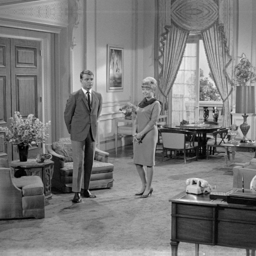}}
\caption{The 12 widely used testing images.}\label{fig111}
\end{center}
\end{figure*}

For the test images, we use two different test datasets for thorough evaluation, one is a test dataset containing 68 natural images from Berkeley segmentation dataset (BSD68)~\cite{roth2009fields} and the other one contains 12 images as shown in Fig.~\ref{fig111}. Note that all those images are widely used for the evaluation of Gaussian denoising methods and they are not included in the training dataset.

In addition to gray image denoising, we also train the blind color image denoising model referred to as CDnCNN-B. We use color version of the BSD68 dataset for testing and
the remaining 432 color images from Berkeley segmentation dataset are adopted as the training images. The noise levels are also set into the range of $[0, 55]$ and $128\times3,000$ patches of size 50$\times$50 are cropped to train the model.

{To learn a single model for the three general image denoising tasks}, as in~\cite{kim2015accurate}, we use a dataset which consists of 91 images from~\cite{yang2010image} and 200 training images from the Berkeley segmentation dataset. The noisy image is generated by adding Gaussian noise with a certain noise level from the range of $[0, 55]$. The SISR input is generated by first bicubic downsampling and then bicubic upsampling the high-resolution image with downscaling factors 2, 3 and 4. The JPEG deblocking input is generated by compressing the image with a quality factor ranging from 5 to 99 using the MATLAB JPEG encoder. All these images are treated as the inputs to a single DnCNN model. Totally, we generate 128$\times$8,000 image patch (the size is 50 $\times$ 50) pairs for training. Rotation/flip based operations on the patch pairs are used during mini-batch learning. The parameters are initialized with DnCNN-B. We refer to our single DnCNN model for these three general image denoising tasks as DnCNN-3. To test DnCNN-3, we adopt different test set for each task, and the detailed description will be given in Section~\ref{sec: expGID}.

\begin{table*}[!htbp]
\caption{The average PSNR(dB) results of different methods on the BSD68 dataset. The best results are highlighted in bold.}
\center
\begin{tabular}{|c|c|c|c|c|c|c|c|c|}
  \hline
  Methods & \;BM3D\;&  \;WNNM\;&\; EPLL\;& \;MLP\;  & \;CSF\; & \;TNRD\; & \;DnCNN-S\; & \;DnCNN-B\;\\ \hline
  $\sigma = 15$ & 31.07 &   31.37&31.21 & -  & 31.24 & 31.42 & \textbf{31.73}   &31.61\\\hline
  $\sigma = 25$ & 28.57 &  28.83 &28.68 & 28.96 &  28.74 & 28.92 & \textbf{29.23}& 29.16\\\hline
  $\sigma = 50$ & 25.62 &  25.87 &25.67 & 26.03 &  - & 25.97 & \textbf{26.23} &\textbf{26.23}\\
  \hline
\end{tabular}
\label{table1}
\end{table*}

\subsubsection{Parameter Setting and Network Training}

In order to capture enough spatial information for denoising, we set the network depth to 17 for DnCNN-S and 20 for DnCNN-B and DnCNN-3. The loss function in Eqn.~\eqref{eq:loss} is adopted to learn the residual mapping $\mathcal{R}(\mathbf{y})$ for predicting the residual $\mathbf{v}$.
We initialize the weights by the method in~\cite{he2015delving} and use SGD with weight decay of 0.0001, a momentum of 0.9 and a mini-batch size of 128.
We train 50 epochs for our DnCNN models.
The learning rate was decayed exponentially from $1e-1$ to $1e-4$ for the 50 epochs.

We use the MatConvNet package~\cite{vedaldi2015matconvnet} to train the proposed DnCNN models.
Unless otherwise specified, all the experiments are carried out in the Matlab (R2015b) environment running on a PC
with Intel(R) Core(TM) i7-5820K CPU 3.30GHz and an Nvidia Titan X GPU. It takes about 6 hours, one day and three days to train DnCNN-S, DnCNN-B/CDnCNN-B and DnCNN-3 on GPU, respectively.

\begin{figure*}[htbp]
\begin{center}
\subfigure[Noisy / 14.76dB]
{\includegraphics[width=0.244\textwidth]{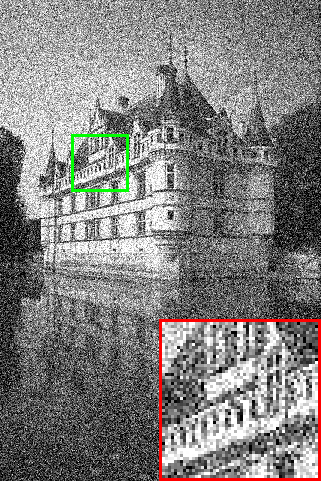}}
\subfigure[BM3D / 26.21dB]
{\includegraphics[width=0.244\textwidth]{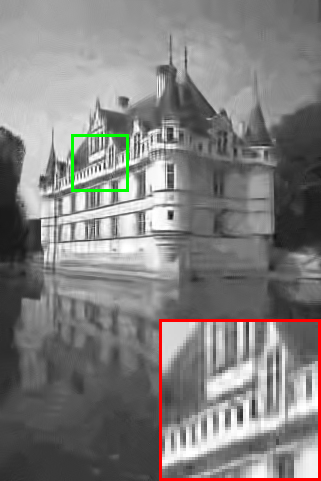}}
\subfigure[WNNM / 26.51dB]
{\includegraphics[width=0.244\textwidth]{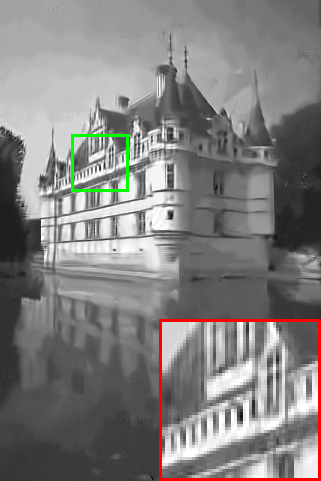}}
\subfigure[EPLL / 26.36dB]
{\includegraphics[width=0.244\textwidth]{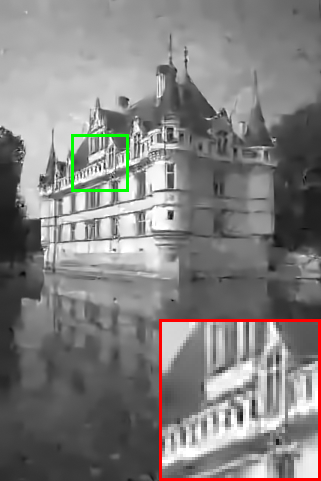}}
\subfigure[MLP / 26.54dB]
{\includegraphics[width=0.244\textwidth]{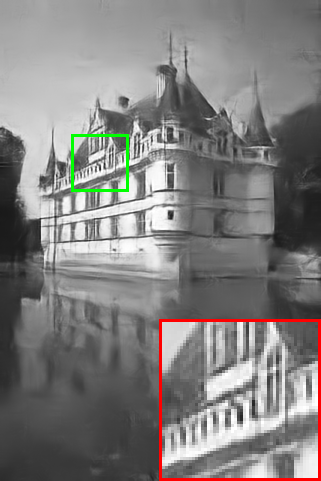}}
\subfigure[TNRD / 26.59dB]
{\includegraphics[width=0.244\textwidth]{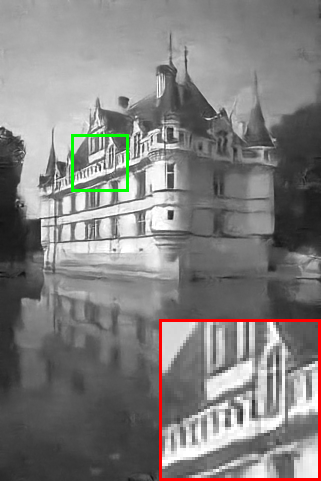}}
\subfigure[DnCNN-S / 26.90dB]
{\includegraphics[width=0.244\textwidth]{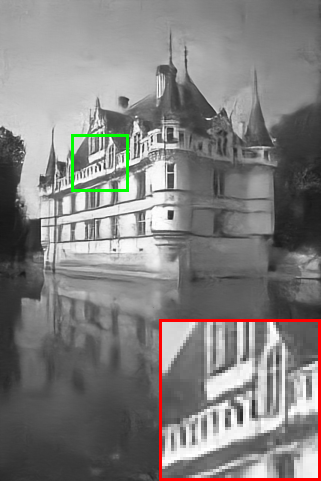}}
\subfigure[DnCNN-B / 26.92dB]
{\includegraphics[width=0.244\textwidth]{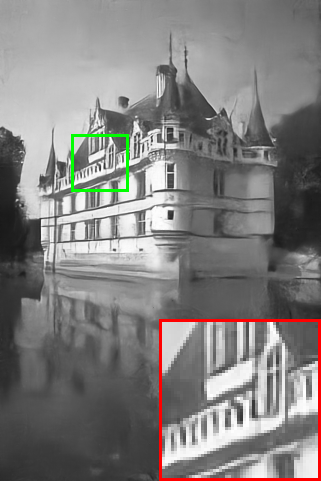}}%
\caption{Denoising results of one image from BSD68 with noise level 50.}\label{fig31}
\end{center}
\end{figure*}

\begin{figure*}[!htbp]
\begin{center}
\subfigure[\scriptsize Noisy / 15.00dB]
{\includegraphics[width=0.244\textwidth]{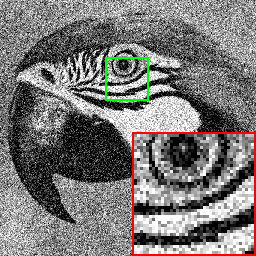}}
\subfigure[\scriptsize BM3D / 25.90dB]
{\includegraphics[width=0.244\textwidth]{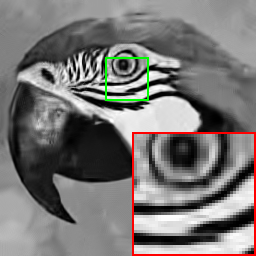}}
\subfigure[\scriptsize WNNM / 26.14dB]
{\includegraphics[width=0.244\textwidth]{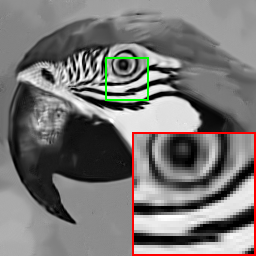}}
\subfigure[\scriptsize EPLL / 25.95dB]
{\includegraphics[width=0.244\textwidth]{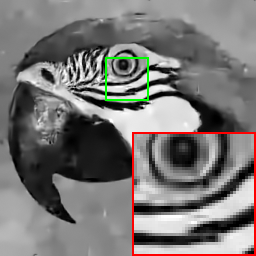}}
\subfigure[\scriptsize MLP / 26.12dB]
{\includegraphics[width=0.244\textwidth]{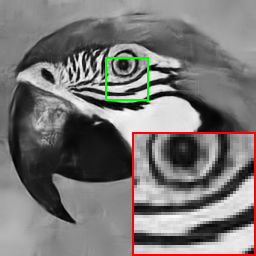}}
\subfigure[\scriptsize TNRD / 26.16dB]
{\includegraphics[width=0.244\textwidth]{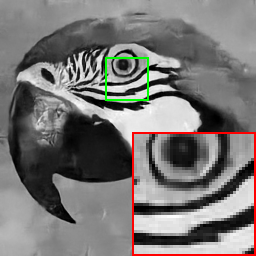}}
\subfigure[\scriptsize DnCNN-S / 26.48dB]
{\includegraphics[width=0.244\textwidth]{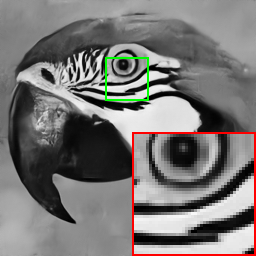}}
\subfigure[\scriptsize DnCNN-B / 26.48dB]
{\includegraphics[width=0.244\textwidth]{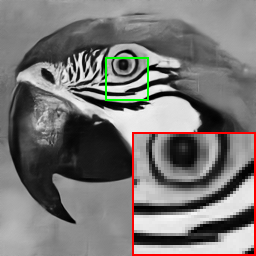}}
\caption{Denoising results of the image ``parrot'' with noise level 50.}\label{fig33}
\end{center}
\end{figure*}

\begin{table*}[htbp]
\caption{The PSNR(dB) results of different  methods on 12 widely used testing images.}
\center
\begin{tabular}{|l|c|c|c|c|c|c|c|c|c|c|c|c|c|}
  \hline
  Images & C.man & House & Peppers & Starfish & Monar. & Airpl. & Parrot & Lena & Barbara & Boat & Man & Couple & Average \\ \hline \hline
    Noise Level & \multicolumn{13}{c|}{$\sigma=15$}   \\ \hline
    BM3D~\cite{dabov2007image} & 31.91 & 34.93 & 32.69 & 31.14 & 31.85 & 31.07 & 31.37 & 34.26 & 33.10 & 32.13 & 31.92 & 32.10 & 32.372  \\\hline
    WNNM~\cite{gu2014weighted}& 32.17 & \textbf{35.13} & 32.99 & 31.82 & 32.71 & 31.39 & 31.62 & 34.27 & \textbf{33.60} & 32.27 & 32.11 & 32.17 &  32.696  \\\hline
     EPLL~\cite{zoran2011learning}& 31.85 & 34.17 & 32.64 & 31.13 & 32.10 & 31.19 & 31.42 & 33.92 & 31.38 & 31.93 & 32.00 & 31.93 & 32.138   \\\hline
     CSF~\cite{schmidt2014shrinkage}& 31.95 & 34.39 & 32.85 & 31.55 & 32.33 & 31.33 & 31.37 & 34.06 & 31.92 & 32.01 & 32.08 & 31.98 &  32.318  \\\hline
     TNRD~\cite{chen2015trainable}& 32.19 & 34.53 & 33.04 & 31.75 & 32.56 & 31.46 & 31.63 & 34.24 & 32.13 & 32.14 & 32.23 & 32.11 & 32.502  \\\hline
     DnCNN-S& \textbf{32.61} & 34.97 & \textbf{33.30} & \textbf{32.20} & \textbf{33.09} & \textbf{31.70} & \textbf{31.83} & \textbf{34.62} & 32.64 & \textbf{32.42} & \textbf{32.46} & \textbf{32.47} &  \textbf{32.859}  \\\hline
     DnCNN-B& 32.10 & 34.93 & 33.15 & 32.02 & 32.94 & 31.56 & 31.63 & 34.56 & 32.09 & 32.35 & 32.41 & 32.41 &  32.680  \\\hline\hline

     Noise Level& \multicolumn{13}{c|}{$\sigma=25$}   \\ \hline
    BM3D~\cite{dabov2007image} & 29.45 & 32.85 & 30.16 & 28.56 & 29.25 & 28.42 & 28.93 & 32.07 & 30.71 & 29.90 & 29.61 & 29.71 &  29.969  \\\hline
     WNNM~\cite{gu2014weighted}& 29.64 & \textbf{33.22} & 30.42 & 29.03 & 29.84 & 28.69 & 29.15 & 32.24 & \textbf{31.24} & 30.03 & 29.76 & 29.82 & 30.257   \\\hline
     EPLL~\cite{zoran2011learning}& 29.26 & 32.17 & 30.17 & 28.51 & 29.39 & 28.61 & 28.95 & 31.73 & 28.61 & 29.74 & 29.66 & 29.53 & 29.692   \\\hline
     MLP~\cite{burger2012image}& 29.61 & 32.56 & 30.30 & 28.82 & 29.61 & 28.82 & 29.25 & 32.25 & 29.54 & 29.97 & 29.88 & 29.73 & 30.027   \\\hline
     CSF~\cite{schmidt2014shrinkage}& 29.48 & 32.39 & 30.32 & 28.80 & 29.62 & 28.72 & 28.90 & 31.79 & 29.03 & 29.76 & 29.71 & 29.53 &  29.837  \\\hline
     TNRD~\cite{chen2015trainable}& 29.72 & 32.53 & 30.57 & 29.02 & 29.85 & 28.88 & 29.18 & 32.00 & 29.41 & 29.91 & 29.87 & 29.71 & 30.055    \\\hline
     DnCNN-S& \textbf{30.18} & 33.06 & \textbf{30.87} & \textbf{29.41} & \textbf{30.28} & \textbf{29.13} & \textbf{29.43} & \textbf{32.44} & 30.00 & \textbf{30.21} & \textbf{30.10 }& \textbf{30.12} &  \textbf{30.436}  \\\hline
     DnCNN-B& 29.94 & 33.05 & 30.84 & 29.34 &30.25 & 29.09 & 29.35 & 32.42 & 29.69 & 30.20 & 30.09 & 30.10 &  30.362  \\\hline \hline

     Noise Level& \multicolumn{13}{c|}{$\sigma=50$}   \\ \hline
    BM3D~\cite{dabov2007image} & 26.13 & 29.69 & 26.68 & 25.04 & 25.82 & 25.10 & 25.90 & 29.05 & 27.22 & 26.78 & 26.81 & 26.46 &  26.722  \\\hline
    WNNM~\cite{gu2014weighted}& 26.45 & \textbf{30.33} & 26.95 & 25.44 & 26.32 & 25.42 & 26.14 & 29.25 & \textbf{27.79} & 26.97 & 26.94 & 26.64 & 27.052   \\\hline
     EPLL~\cite{zoran2011learning}& 26.10 & 29.12 & 26.80 & 25.12 & 25.94 & 25.31 & 25.95 & 28.68 & 24.83 & 26.74 & 26.79 & 26.30 & 26.471   \\\hline
     MLP~\cite{burger2012image}& 26.37 & 29.64 & 26.68 & 25.43 & 26.26 & 25.56 & 26.12 & 29.32 & 25.24 & 27.03 & 27.06 & 26.67 &  26.783  \\\hline
     TNRD~\cite{chen2015trainable}& 26.62 & 29.48 & 27.10 & 25.42 & 26.31 & 25.59 & 26.16 & 28.93 & 25.70 & 26.94 & 26.98 & 26.50 & 26.812   \\\hline
     DnCNN-S& \textbf{27.03} & 30.00 & 27.32 & 25.70 & 26.78 & 25.87 & \textbf{26.48} & \textbf{29.39} & 26.22 & 27.20 & \textbf{27.24} & 26.90 &  27.178  \\\hline
     DnCNN-B & \textbf{27.03} & 30.02 & \textbf{27.39} & \textbf{25.72} & \textbf{26.83} & \textbf{25.89} & \textbf{26.48} & 29.38 & 26.38 & \textbf{27.23} & 27.23 & \textbf{26.91} &  \textbf{27.206}  \\\hline
\end{tabular}
\label{table888}
\end{table*}

\subsection{Compared Methods}

We compare the proposed DnCNN method with several state-of-the-art denoising methods, including two non-local similarity based methods (i.e., BM3D~\cite{dabov2007image} and WNNM~\cite{gu2014weighted}), one generative method (i.e., EPLL~\cite{zoran2011learning}), three discriminative training based methods (i.e., MLP~\cite{burger2012image}, CSF~\cite{schmidt2014shrinkage} and TNRD~\cite{chen2015trainable}). Note that CSF and TNRD are highly efficient by GPU implementation while offering good image quality. The implementation codes are downloaded from the authors' websites and the default parameter settings are used in our experiments.  The testing code of our DnCNN models can
be downloaded at \url{https://github.com/cszn/DnCNN}.

\subsection{Quantitative and Qualitative Evaluation}

The average PSNR results of different methods on the BSD68 dataset are shown in Table~\ref{table1}.
As one can see, both DnCNN-S and DnCNN-B can achieve the best PSNR results than the competing methods.
Compared to the benchmark BM3D, the methods MLP and TNRD have a notable PSNR gain of about 0.35dB.
According to~\cite{levin2011natural,levin2012patch}, few methods can outperform BM3D by more than 0.3dB on average. In contrast, our DnCNN-S model outperforms BM3D by 0.6dB on all the three noise levels. Particularly, even with a single model without known noise level, our DnCNN-B can still outperform the competing methods which is trained for the known specific noise level.
It should be noted that both DnCNN-S and DnCNN-B outperform BM3D by about 0.6dB when $\sigma=50$, which is very close to the estimated PSNR bound over BM3D (0.7dB) in~\cite{levin2012patch}.

Table~\ref{table888} lists the PSNR results of different methods on the 12 test images in Fig. \ref{fig111}. The best PSNR result for each image with each noise level is highlighted in bold.
It can be seen that the proposed DnCNN-S yields the highest PSNR on most of the images. Specifically, DnCNN-S outperforms the competing methods by 0.2dB to 0.6dB on most of the images and fails to achieve the best results on only two images ``House'' and ``Barbara'', which are dominated by repetitive structures.
This result is consistent with the findings in~\cite{burger2013learning}: non-local means based methods are usually better on images with regular and repetitive structures whereas discriminative training based methods generally produce better results on images with irregular textures. Actually, this is intuitively reasonable because images with regular and repetitive structures meet well with the non-local similarity prior; conversely, images with irregular textures would weaken the advantages of such specific prior, thus leading to poor results.

Figs.~\ref{fig31}-\ref{fig33} illustrate the visual results of different methods.
It can be seen that BM3D, WNNM, EPLL and MLP tend to produce over-smooth edges and textures.
While preserving sharp edges and fine details, TNRD is likely to generate artifacts in the smooth region.
In contrast, DnCNN-S and DnCNN-B can not only recover sharp edges and fine details but also yield visually pleasant results in the smooth region.

\begin{figure*}[htbp]
\begin{center}
\subfigure[Ground-truth]
{\includegraphics[width=0.244\textwidth]{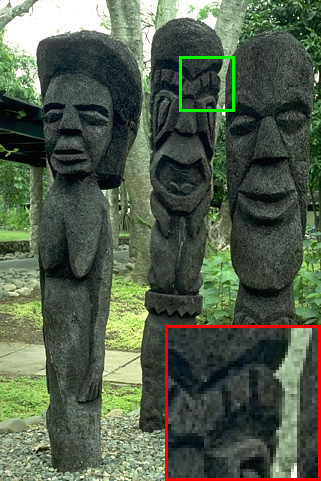}}
\subfigure[Noisy / 17.25dB]
{\includegraphics[width=0.244\textwidth]{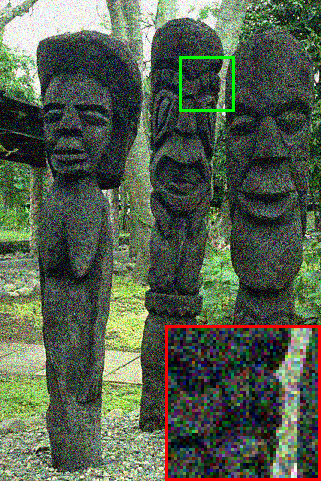}}
\subfigure[CBM3D / 25.93dB]
{\includegraphics[width=0.244\textwidth]{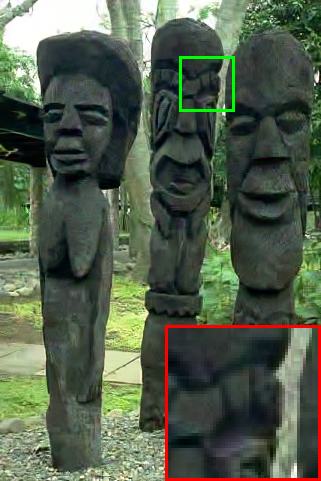}}
\subfigure[CDnCNN-B / 26.58dB]
{\includegraphics[width=0.244\textwidth]{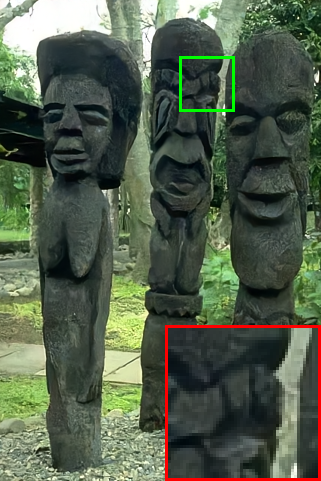}}
\caption{Color image denoising results of one image from the DSD68 dataset with noise level 35.}\label{fig_c1}
\end{center}
\end{figure*}

\begin{figure*}[htbp]
\begin{center}
\subfigure[Ground-truth]
{\includegraphics[width=0.244\textwidth]{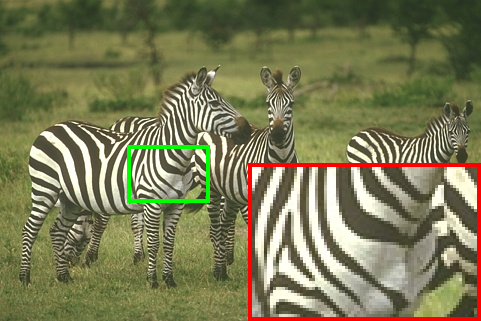}}
\subfigure[Noisy / 15.07dB]
{\includegraphics[width=0.244\textwidth]{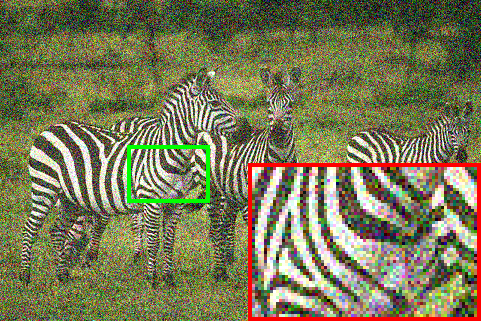}}
\subfigure[CBM3D / 26.97dB]
{\includegraphics[width=0.244\textwidth]{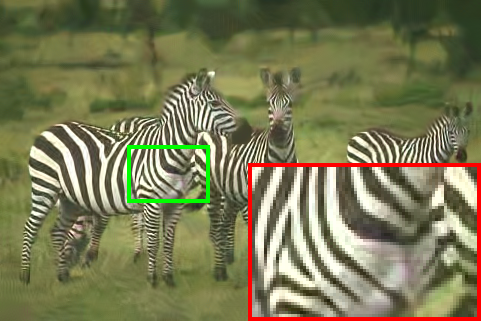}}
\subfigure[CDnCNN-B / 27.87dB]
{\includegraphics[width=0.244\textwidth]{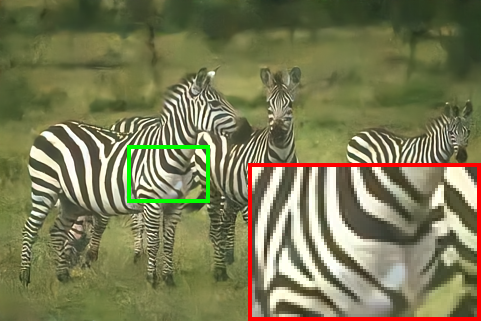}}
\caption{Color image denoising results of one image from the DSD68 dataset with noise level 45.}\label{fig_c2}
\end{center}
\end{figure*}

For color image denoising, the visual comparisons between CDnCNN-B and the benchmark CBM3D are shown in Figs.~\ref{fig_c1}-\ref{fig_c2}. One can see that
CBM3D generates false color artifacts in some regions whereas CDnCNN-B can recover images with more natural color.
In addition, CDnCNN-B can generate images with more details and sharper edges than CBM3D.

\begin{figure}[htbp]
  \centering
  \includegraphics[width=0.49\textwidth]{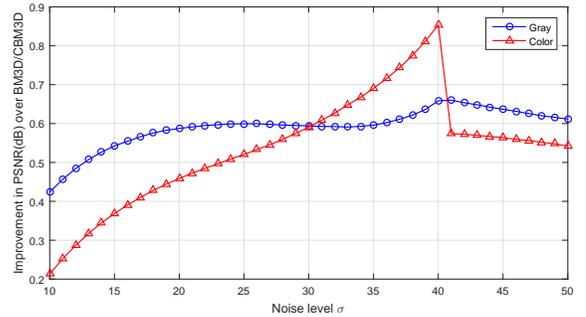}\\
  \caption{Average PSNR improvement over BM3D/CBM3D with respect to different noise levels by our DnCNN-B/CDnCNN-B model. The results are evaluated on the gray/color BSD68 dataset.}\label{fig8}
\end{figure}

Fig.~\ref{fig8} shows the average PSNR improvement over BM3D/CBM3D with respect to different noise levels by DnCNN-B/CDnCNN-B model. It can be seen that our DnCNN-B/CDnCNN-B models consistently outperform BM3D/CBM3D by a large margin on a wide range of noise levels. This experimental result demonstrates the feasibility of training a single DnCNN-B model for handling blind Gaussian denoising within a wide range of noise levels.

\subsection{Run Time}

In addition to visual quality, another important aspect for an image restoration method is the testing speed.
Table~\ref{table2} shows the run times of different methods for denoising images of sizes $256 \times 256$, $512 \times 512$ and $1024 \times 1024$ with noise level 25.
Since CSF, TNRD and our DnCNN methods are well-suited for parallel computation on GPU, we also give the corresponding run times on GPU.
We use the Nvidia cuDNN-v5 deep learning library to accelerate the GPU computation of the proposed DnCNN.
As in~\cite{chen2015trainable}, we do not count the memory transfer time between CPU and GPU.
It can be seen that the proposed DnCNN can have a relatively high speed on CPU and it is faster than two discriminative models, MLP and CSF.
Though it is slower than BM3D and TNRD, by taking the image quality improvement into consideration, our DnCNN is still very competitive in CPU implementation. For the GPU time, the proposed DnCNN achieves very appealing computational efficiency, e.g., it can denoise an image of size $512\times512$ in 60ms with unknown noise level, which is a distinct advantage over TNRD.

\begin{table*}[!htbp]
  \centering
  \caption{Run time (in seconds) of different methods on images of size $256 \times 256$, $512 \times 512$ and $1024 \times 1024$ with noise level 25. For CSF, TNRD and our proposed DnCNN, we give the run times on CPU (left) and GPU (right).
  It is also worth noting that since the run time on GPU varies greatly with respect to GPU and GPU-accelerated library, it is hard to make a fair comparison between CSF, TNRD and our proposed DnCNN. Therefore, we just copy the run times of CSF and TNRD on GPU from the original papers.}
  \begin{tabular}{|c|c|c|c|c|c|c|c|c|}
    \hline
    Methods & BM3D  & WNNM & EPLL& MLP & CSF & TNRD & DnCNN-S & DnCNN-B \\ \hline\hline
    256$\times$256 & 0.65   & 203.1& 25.4 & 1.42& 2.11 / ~~-~~ & 0.45 / 0.010 &  0.74 / 0.014&  0.90 / 0.016\\ \hline
    512$\times$512 & 2.85  & 773.2& 45.5 & 5.51 & 5.67 / 0.92 & 1.33 / 0.032 &  3.41 / 0.051 &  4.11 / 0.060\\ \hline
    1024$\times$1024 & 11.89  & 2536.4&  422.1 & 19.4& 40.8 / 1.72 & 4.61 / 0.116 & 12.1 / 0.200 & 14.1 / 0.235\\
    \hline
  \end{tabular}
  \label{table2}
\end{table*}

\subsection{{Experiments on Learning a Single Model for Three General Image Denoising Tasks}}\label{sec: expGID}

\begin{table}[!htbp]\scriptsize
\caption{Average PSNR(dB)/SSIM results of different methods for Gaussian denoising with noise level 15, 25 and 50 on BSD68 dataset, single image super-resolution with upscaling factors 2, 3 and 4 on Set5, Set14, BSD100 and Urban100 datasets, JPEG image deblocking with quality factors 10, 20, 30 and 40 on  Classic5 and LIVE1 datasets. The best results are highlighted in bold.} 
\center
\begin{tabular}{|l|c|c|c|c|}

 \hline \multicolumn{5}{|c|}{\textbf{Gaussian Denoising}}    \\ \hline
  \multirow{2}{*}{Dataset} & Noise & BM3D & TNRD & DnCNN-3  \\
  \cline{3-5}
  & Level & PSNR / SSIM & PSNR / SSIM & PSNR / SSIM  \\ \hline
     &15 & 31.08 / 0.8722 & 31.42 / \textbf{0.8826}&\textbf{31.46} / \textbf{0.8826}  \\
 BSD68 &25 & 28.57 / 0.8017 & 28.92 / 0.8157 &\textbf{29.02} / \textbf{0.8190}  \\
  &50 & 25.62 / 0.6869 &  25.97 / 0.7029 &\textbf{26.10} / \textbf{0.7076}  \\ \hline \hline

   \multicolumn{5}{|c|}{\textbf{Single Image Super-Resolution}}    \\ \hline

 \multirow{2}{*}{Dataset} & Upscaling & TNRD & VDSR & DnCNN-3   \\
 \cline{3-5}
  & Factor & PSNR / SSIM & PSNR / SSIM & PSNR / SSIM  \\ \hline
  &2 & 36.86 / 0.9556 &  37.56 / \textbf{0.9591} &\textbf{37.58} / 0.9590   \\
 Set5 & 3 & 33.18 / 0.9152 &  33.67 / 0.9220 &\textbf{33.75} / \textbf{0.9222}  \\
  & 4 & 30.85 / 0.8732 &  31.35 / \textbf{0.8845} &\textbf{31.40} / \textbf{0.8845}  \\\hline

    & 2 & 32.51 / 0.9069 & 33.02 / \textbf{0.9128} &\textbf{33.03} / \textbf{0.9128}  \\
 Set14 &3 & 29.43 / 0.8232 & 29.77 / 0.8318 &\textbf{29.81} / \textbf{0.8321}  \\
  & 4 & 27.66 / 0.7563 &  27.99 / 0.7659 &\textbf{28.04} / \textbf{0.7672}  \\\hline

    &2 & 31.40 / 0.8878 &  31.89 / \textbf{0.8961} & \textbf{31.90} / \textbf{0.8961}   \\
 BSD100 &3 & 28.50 / 0.7881 &  28.82 / 0.7980 &\textbf{28.85} / \textbf{0.7981} \\
  & 4 & 27.00 / 0.7140&  27.28 / \textbf{0.7256} &\textbf{27.29} / 0.7253  \\\hline

    &2 & 29.70 / 0.8994 &   \textbf{30.76} / \textbf{0.9143} &30.74 / 0.9139   \\
Urban100 &3 & 26.42 / 0.8076 &  27.13 / \textbf{0.8283} &\textbf{27.15} / 0.8276 \\
  & 4 & 24.61 / 0.7291 &  25.17 / \textbf{0.7528} &\textbf{25.20} / 0.7521  \\\hline \hline

   \multicolumn{5}{|c|}{\textbf{JPEG Image Deblocking}}    \\ \hline

 \multirow{2}{*}{Dataset} & Quality & AR-CNN & TNRD & DnCNN-3  \\
 \cline{3-5}
  & Factor & PSNR / SSIM & PSNR / SSIM & PSNR / SSIM  \\ \hline

 \multirow{4}{*}{Classic5} &10  &  29.03 / 0.7929&29.28 / 0.7992  & \textbf{29.40} / \textbf{0.8026}    \\
  &20 &  31.15 / 0.8517 &31.47 / 0.8576  & \textbf{31.63} / \textbf{0.8610}    \\
  &30 &  32.51 / 0.8806  &32.78 / 0.8837  & \textbf{32.91} / \textbf{0.8861}    \\
&40  &  33.34 / 0.8953 &- & \textbf{33.77} / \textbf{0.9003}  \\\hline

 \multirow{4}{*}{LIVE1}  &10 & 28.96 / 0.8076 &29.15 / 0.8111  & \textbf{29.19} / \textbf{0.8123}   \\
  &20 & 31.29 / 0.8733  &31.46 / 0.8769  & \textbf{31.59} / \textbf{0.8802}  \\
  &30  &  32.67 / 0.9043 &32.84 / 0.9059  & \textbf{32.98} / \textbf{0.9090}  \\
&40 &  33.63 / 0.9198  &- & \textbf{33.96} / \textbf{0.9247}   \\\hline

\end{tabular}
\label{tablem1}
\end{table}

In order to further show the capacity of the proposed DnCNN model, a single DnCNN-3 model is trained for three general image denoising tasks, including blind Gaussian denoising, SISR and JPEG image deblocking. To the best of our knowledge, none of the existing methods have been reported for handling these three tasks with only a single model. Therefore, for each task, we compare DnCNN-3 with the specific state-of-the-art methods. In the following, we describe the compared methods and the test dataset for each task:

\begin{itemize}
  \item For Gaussian denoising, we use the state-of-the-art BM3D and TNRD for comparison. The BSD68 dataset are used for testing the performance. For BM3D and TNRD, we assume that the noise level is known.
  \item For SISR, we consider two state-of-the-art methods, i.e., TNRD and VDSR~\cite{kim2015accurate}.
    TNRD trained a specific model for each upscalling factor while VDSR~\cite{kim2015accurate} trained a single model for all the three upscaling factors (i.e., 2, 3 and 4). We adopt the four testing datasets (i.e., Set5 and Set14, BSD100 and Urban100~\cite{huang2015single}) used in~\cite{kim2015accurate}.
   \item For JPEG image deblocking, our DnCNN-3 is compared with two state-of-the-art methods, i.e., AR-CNN~\cite{dong2015compression} and TNRD~\cite{chen2015trainable}. The AR-CNN method trained four specific models
    for the JPEG quality factors 10, 20, 30 and 40, respectively. For TNRD, three models for JPEG quality factors
    10, 20 and 30 are trained. As in~\cite{dong2015compression}, we adopt the Classic5 and LIVE1 as test datasets.
\end{itemize}

Table~\ref{tablem1} lists the average PSNR and SSIM results of different methods for different general image denoising tasks.
As one can see, even we train a single DnCNN-3 model for the three different tasks, it still outperforms the nonblind TNRD and BM3D for Gaussian denoising. For SISR, it surpasses TNRD by a large margin and is on par with VDSR. For JPEG image deblocking, DnCNN-3 outperforms AR-CNN by about 0.3dB in PSNR and has about 0.1dB PSNR gain over TNRD on all the quality factors.

\begin{figure*}[!htbp]
\begin{center}
\subfigure[Ground-truth]
{\includegraphics[width=0.244\textwidth]{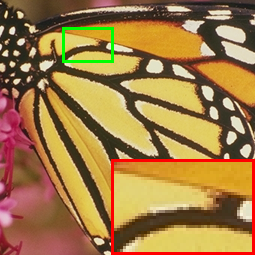}}
\subfigure[TNRD / 28.91dB]
{\includegraphics[width=0.244\textwidth]{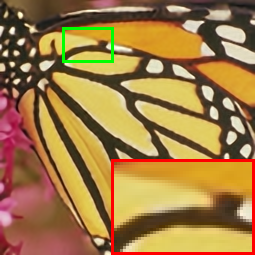}}
\subfigure[VDSR / 29.95dB]
{\includegraphics[width=0.244\textwidth]{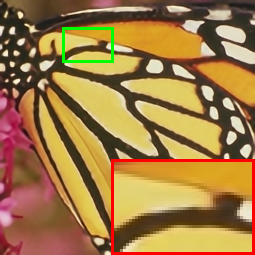}}
\subfigure[DnCNN-3 / 30.02dB]
{\includegraphics[width=0.244\textwidth]{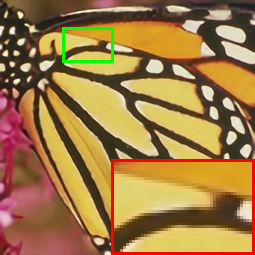}}
\caption{Single image super-resolution results of ``butterfly'' from Set5 dataset with upscaling factor 3.}\label{figSR31}
\end{center}
\end{figure*}

\begin{figure*}[!htbp]
\begin{center}
\subfigure[Ground-truth]
{\includegraphics[width=0.244\textwidth]{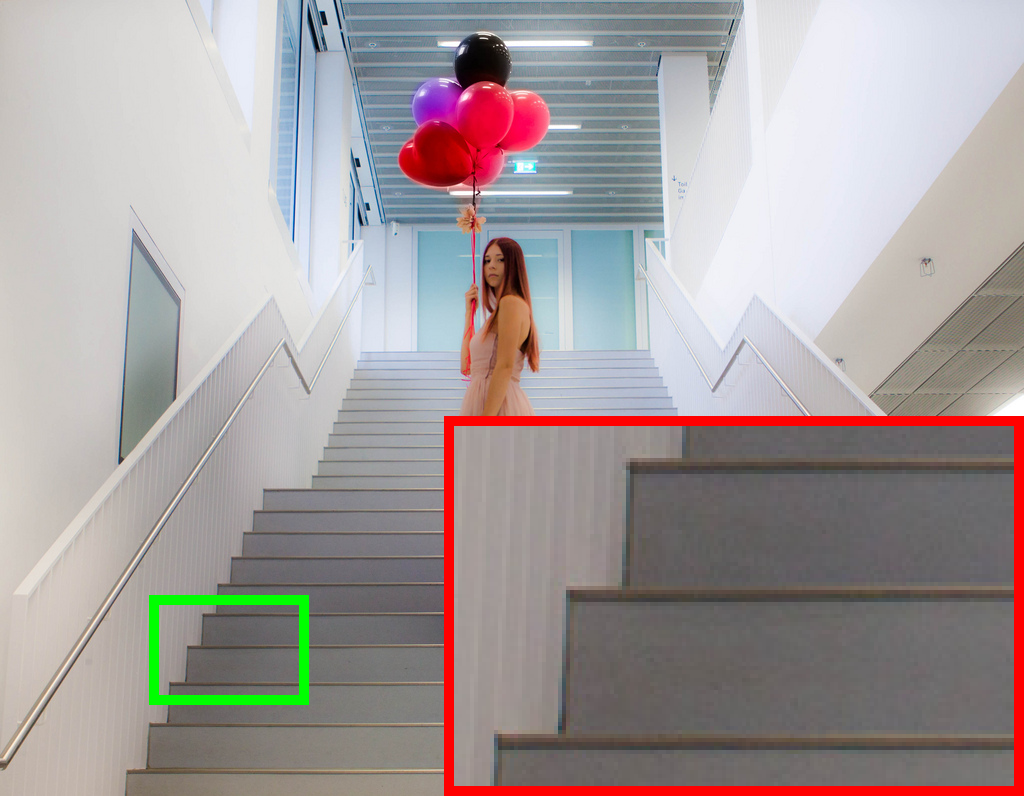}}
\subfigure[TNRD / 32.00dB]
{\includegraphics[width=0.244\textwidth]{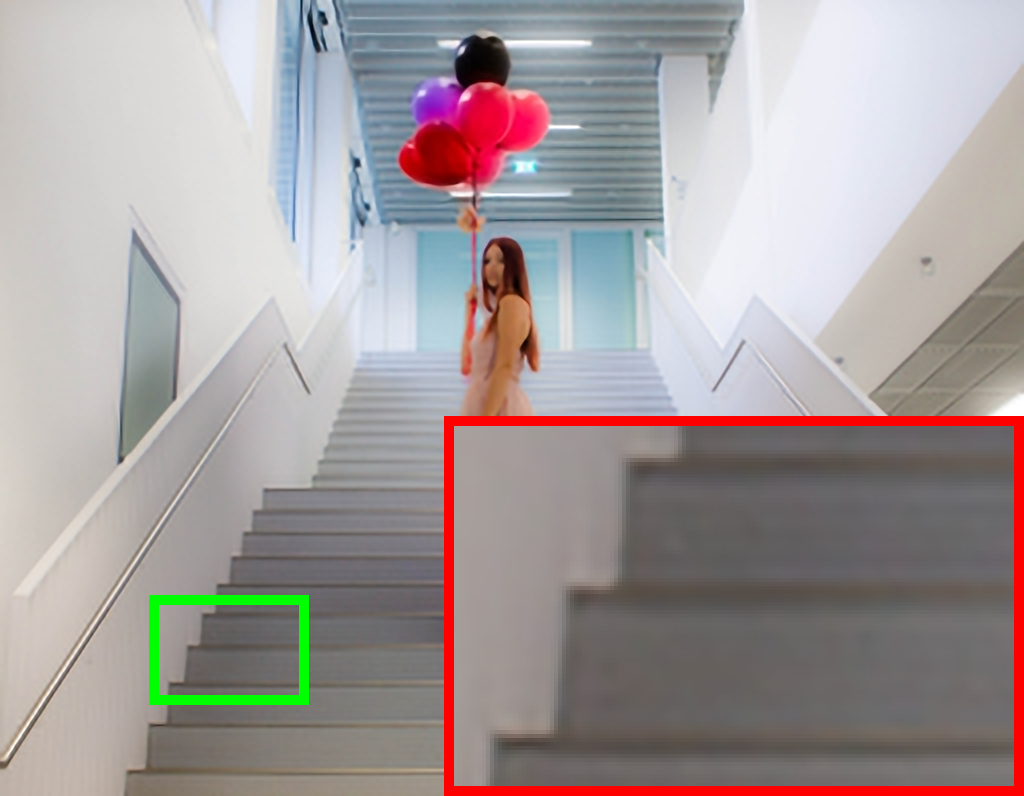}}
\subfigure[VDSR / 32.58dB]
{\includegraphics[width=0.244\textwidth]{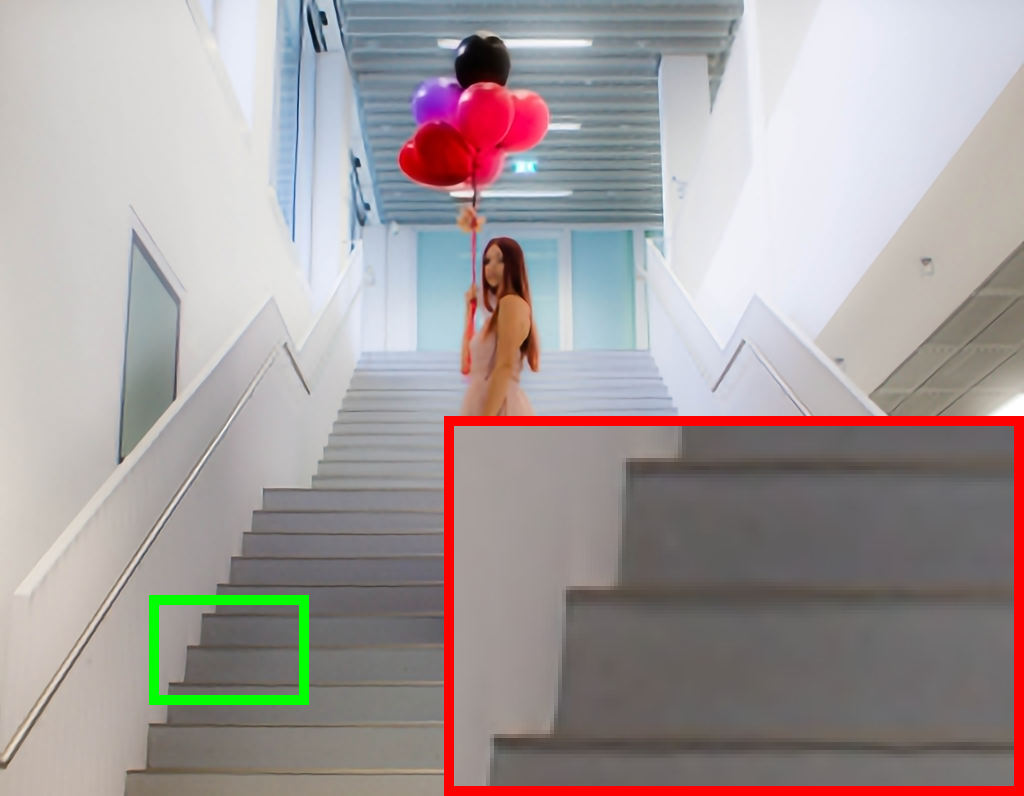}}
\subfigure[DnCNN-3 / 32.73dB]
{\includegraphics[width=0.244\textwidth]{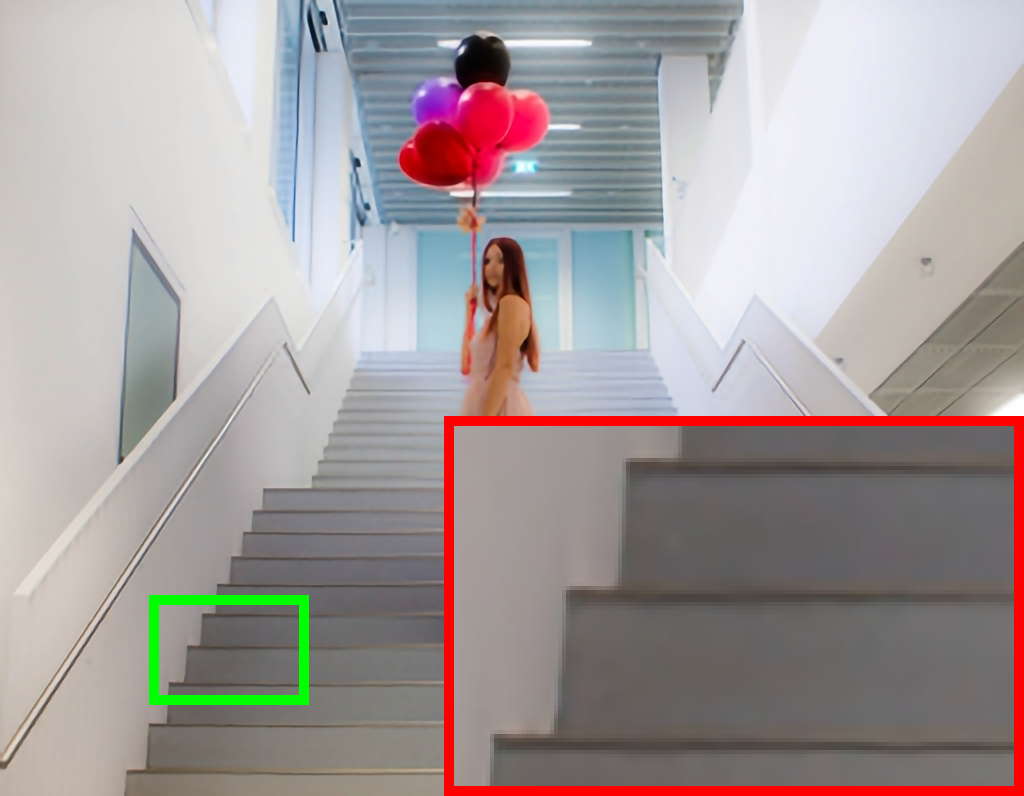}}
\caption{Single image super-resolution results of one image from  Urban100 dataset with upscaling factor 4.}\label{figSR32}
\end{center}
\end{figure*}

\begin{figure*}[!htbp]
\begin{center}
\subfigure[JPEG / 28.10dB]
{\includegraphics[width=0.245\textwidth]{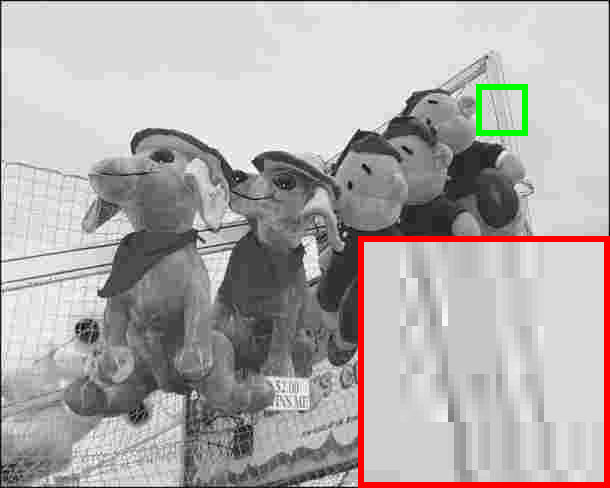}}
\subfigure[AR-CNN / 28.85dB]
{\includegraphics[width=0.245\textwidth]{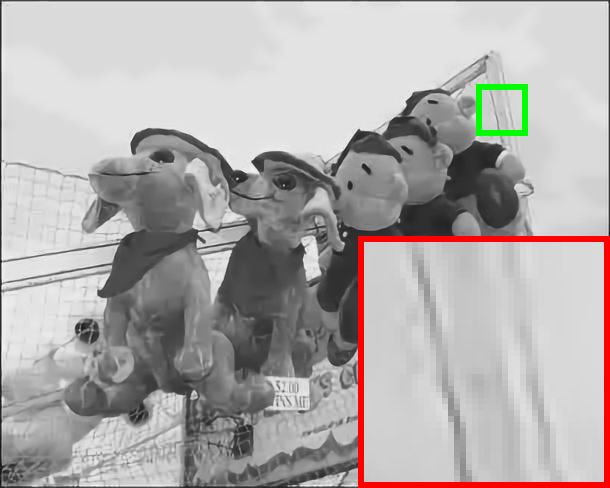}}
\subfigure[TNRD / 29.54dB]
{\includegraphics[width=0.245\textwidth]{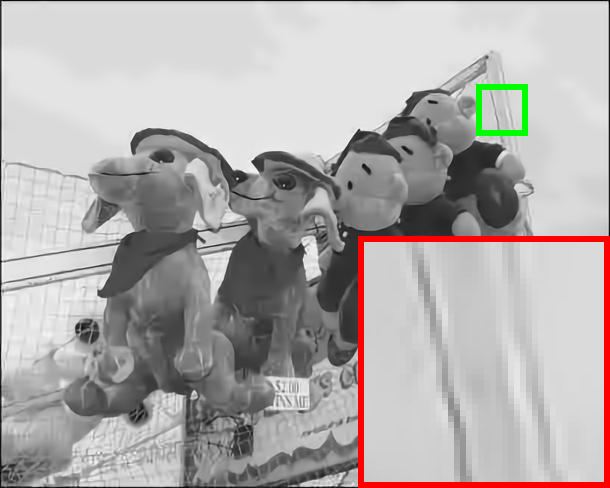}}
\subfigure[DnCNN-3 / 29.70dB]
{\includegraphics[width=0.245\textwidth]{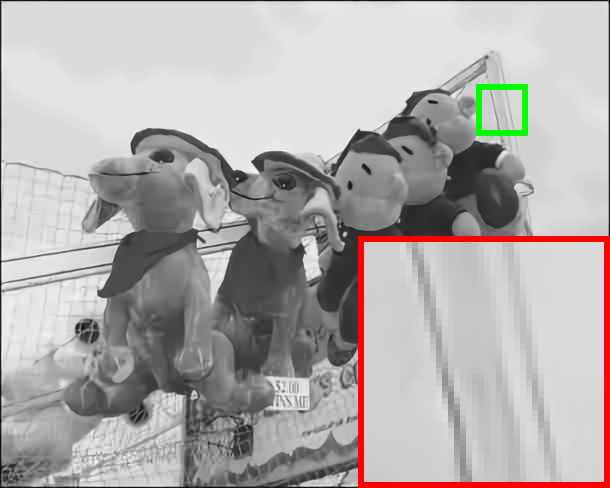}}
\caption{JPEG image deblocking results of ``Carnivaldolls'' from LIVE1 dataset with quality factor 10.}\label{figDB3}
\end{center}
\end{figure*}

\begin{figure*}[!htbp]
\begin{center}
\subfigure[Input Image]
{\includegraphics[width=0.32\textwidth]{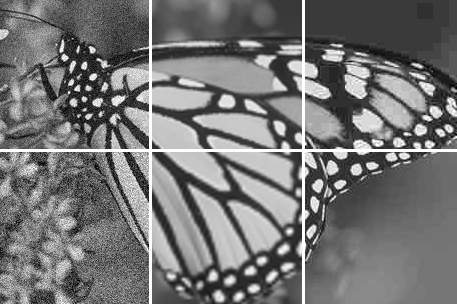}}
\subfigure[Output Residual Image]
{\includegraphics[width=0.32\textwidth]{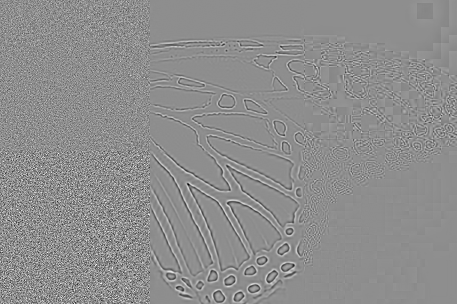}}
\subfigure[Restored Image]
{\includegraphics[width=0.32\textwidth]{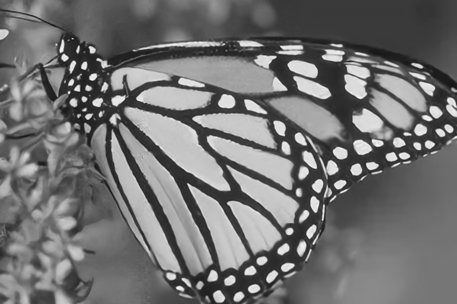}}
\caption{An example to show the capacity of our proposed model for three different tasks. The input image is composed by noisy images with noise level 15 (upper left) and 25 (lower left), bicubically interpolated low-resolution images with upscaling factor 2 (upper middle) and  3 (lower middle), JPEG images with quality factor 10 (upper right) and 30 (lower right). Note that the white lines in the input image are just used for distinguishing the six regions, and the residual image is normalized into the range of $[0, 1]$ for visualization. Even the input image is corrupted with different distortions in different regions, the restored image looks natural and does not have obvious artifacts.}\label{fig03}
\end{center}
\end{figure*}

Fig.~\ref{figSR31} and Fig.~\ref{figSR32} show the visual comparisons of different methods for SISR. It can be seen that both DnCNN-3 and VDSR can produce sharp edges and fine details whereas TNRD tend to generate blurred edges and distorted lines. Fig.~\ref{figDB3} shows the JPEG deblocking results of different methods. As one can see, our DnCNN-3 can recover the straight line whereas AR-CNN and TNRD are prone to generate distorted lines. Fig.~\ref{fig03} gives an additional example to show the capacity of the proposed model. We can see that DnCNN-3 can produce visually pleasant output result even the input image is corrupted by several distortions with different levels in different regions.

\section{Conclusion}\label{sec:conclusion}
In this paper, a deep convolutional neural network was proposed for {image denoising}, where residual learning is adopted to separating noise from noisy observation. The batch normalization and residual learning are integrated to speed up the training process as well as boost the denoising performance. Unlike traditional discriminative models which train specific models for certain noise levels, our single DnCNN model has the capacity to handle the blind Gaussian denoising with unknown noise level. Moreover, we showed the feasibility to train a single DnCNN model to handle three general image denoising tasks, including Gaussian denoising with unknown noise level, single image super-resolution with multiple upscaling factors, and JPEG image deblocking with different quality factors. Extensive experimental results demonstrated that the proposed method not only produces favorable image denoising performance quantitatively and qualitatively but also has promising run time by GPU implementation.



\begin{thebibliography}{10}
\providecommand{\url}[1]{#1}
\csname url@samestyle\endcsname
\providecommand{\newblock}{\relax}
\providecommand{\bibinfo}[2]{#2}
\providecommand{\BIBentrySTDinterwordspacing}{\spaceskip=0pt\relax}
\providecommand{\BIBentryALTinterwordstretchfactor}{4}
\providecommand{\BIBentryALTinterwordspacing}{\spaceskip=\fontdimen2\font plus
\BIBentryALTinterwordstretchfactor\fontdimen3\font minus
  \fontdimen4\font\relax}
\providecommand{\BIBforeignlanguage}[2]{{%
\expandafter\ifx\csname l@#1\endcsname\relax
\typeout{** WARNING: IEEEtran.bst: No hyphenation pattern has been}%
\typeout{** loaded for the language `#1'. Using the pattern for}%
\typeout{** the default language instead.}%
\else
\language=\csname l@#1\endcsname
\fi
#2}}
\providecommand{\BIBdecl}{\relax}
\BIBdecl

\bibitem{buades2005non}
A.~Buades, B.~Coll, and J.-M. Morel, ``A non-local algorithm for image
  denoising,'' in \emph{IEEE Conference on Computer Vision and Pattern
  Recognition}, vol.~2, 2005, pp. 60--65.

\bibitem{dabov2007image}
K.~Dabov, A.~Foi, V.~Katkovnik, and K.~Egiazarian, ``Image denoising by sparse
  3-{D} transform-domain collaborative filtering,'' \emph{IEEE Transactions on
  Image Processing}, vol.~16, no.~8, pp. 2080--2095, 2007.

\bibitem{buades2008nonlocal}
A.~Buades, B.~Coll, and J.-M. Morel, ``Nonlocal image and movie denoising,''
  \emph{International Journal of Computer Vision}, vol.~76, no.~2, pp.
  123--139, 2008.

\bibitem{mairal2009non}
J.~Mairal, F.~Bach, J.~Ponce, G.~Sapiro, and A.~Zisserman, ``Non-local sparse
  models for image restoration,'' in \emph{IEEE International Conference on
  Computer Vision}, 2009, pp. 2272--2279.

\bibitem{elad2006image}
M.~Elad and M.~Aharon, ``Image denoising via sparse and redundant
  representations over learned dictionaries,'' \emph{IEEE Transactions on Image
  Processing}, vol.~15, no.~12, pp. 3736--3745, 2006.

\bibitem{dong2013nonlocally}
W.~Dong, L.~Zhang, G.~Shi, and X.~Li, ``Nonlocally centralized sparse
  representation for image restoration,'' \emph{IEEE Transactions on Image
  Processing}, vol.~22, no.~4, pp. 1620--1630, 2013.

\bibitem{rudin1992nonlinear}
L.~I. Rudin, S.~Osher, and E.~Fatemi, ``Nonlinear total variation based noise
  removal algorithms,'' \emph{Physica D: Nonlinear Phenomena}, vol.~60, no.~1,
  pp. 259--268, 1992.

\bibitem{osher2005iterative}
S.~Osher, M.~Burger, D.~Goldfarb, J.~Xu, and W.~Yin, ``An iterative
  regularization method for total variation-based image restoration,''
  \emph{Multiscale Modeling \& Simulation}, vol.~4, no.~2, pp. 460--489, 2005.

\bibitem{weiss2007makes}
Y.~Weiss and W.~T. Freeman, ``What makes a good model of natural images?'' in
  \emph{IEEE Conference on Computer Vision and Pattern Recognition}, 2007, pp.
  1--8.

\bibitem{lan2006efficient}
X.~Lan, S.~Roth, D.~Huttenlocher, and M.~J. Black, ``Efficient belief
  propagation with learned higher-order {M}arkov random fields,'' in
  \emph{European Conference on Computer Vision}, 2006, pp. 269--282.

\bibitem{li2009markov}
S.~Z. Li, \emph{Markov random field modeling in image analysis}.\hskip 1em plus
  0.5em minus 0.4em\relax Springer Science \& Business Media, 2009.

\bibitem{roth2009fields}
S.~Roth and M.~J. Black, ``Fields of experts,'' \emph{International Journal of
  Computer Vision}, vol.~82, no.~2, pp. 205--229, 2009.

\bibitem{gu2014weighted}
S.~Gu, L.~Zhang, W.~Zuo, and X.~Feng, ``Weighted nuclear norm minimization with
  application to image denoising,'' in \emph{IEEE Conference on Computer Vision
  and Pattern Recognition}, 2014, pp. 2862--2869.

\bibitem{schmidt2014shrinkage}
U.~Schmidt and S.~Roth, ``Shrinkage fields for effective image restoration,''
  in \emph{IEEE Conference on Computer Vision and Pattern Recognition}, 2014,
  pp. 2774--2781.

\bibitem{chen2015learning}
Y.~Chen, W.~Yu, and T.~Pock, ``On learning optimized reaction diffusion
  processes for effective image restoration,'' in \emph{IEEE Conference on
  Computer Vision and Pattern Recognition}, 2015, pp. 5261--5269.

\bibitem{chen2015trainable}
Y.~Chen and T.~Pock, ``Trainable nonlinear reaction diffusion: {A} flexible
  framework for fast and effective image restoration,'' \emph{to appear in IEEE
  transactions on Pattern Analysis and Machine Intelligence}, 2016.

\bibitem{schmidt2013discriminative}
U.~Schmidt, C.~Rother, S.~Nowozin, J.~Jancsary, and S.~Roth, ``Discriminative
  non-blind deblurring,'' in \emph{IEEE Conference on Computer Vision and
  Pattern Recognition}, 2013, pp. 604--611.

\bibitem{Schmidtpami}
U.~Schmidt, J.~Jancsary, S.~Nowozin, S.~Roth, and C.~Rother, ``Cascades of
  regression tree fields for image restoration,'' \emph{IEEE Conference on
  Computer Vision and Pattern Recognition}, vol.~38, no.~4, pp. 677--689, 2016.

\bibitem{simonyan2014very}
K.~Simonyan and A.~Zisserman, ``Very deep convolutional networks for
  large-scale image recognition,'' in \emph{International Conference for
  Learning Representations}, 2015.

\bibitem{krizhevsky2012imagenet}
A.~Krizhevsky, I.~Sutskever, and G.~E. Hinton, ``Imagenet classification with
  deep convolutional neural networks,'' in \emph{Advances in Neural Information
  Processing Systems}, 2012, pp. 1097--1105.

\bibitem{ioffe2015batch}
S.~Ioffe and C.~Szegedy, ``Batch normalization: Accelerating deep network
  training by reducing internal covariate shift,'' in \emph{International
  Conference on Machine Learning}, 2015, pp. 448--456.

\bibitem{he2015deep}
K.~He, X.~Zhang, S.~Ren, and J.~Sun, ``Deep residual learning for image
  recognition,'' in \emph{IEEE Conference on Computer Vision and Pattern
  Recognition}, 2016, pp. 770--778.

\bibitem{jain2009natural}
V.~Jain and S.~Seung, ``Natural image denoising with convolutional networks,''
  in \emph{Advances in Neural Information Processing Systems}, 2009, pp.
  769--776.

\bibitem{burger2012image}
H.~C. Burger, C.~J. Schuler, and S.~Harmeling, ``Image denoising: Can plain
  neural networks compete with {BM3D}?'' in \emph{IEEE Conference on Computer
  Vision and Pattern Recognition}, 2012, pp. 2392--2399.

\bibitem{xie2012image}
J.~Xie, L.~Xu, and E.~Chen, ``Image denoising and inpainting with deep neural
  networks,'' in \emph{Advances in Neural Information Processing Systems},
  2012, pp. 341--349.

\bibitem{szegedy2015googlenet}
C.~Szegedy, W.~Liu, Y.~Jia, P.~Sermanet, S.~Reed, D.~Anguelov, D.~Erhan,
  V.~Vanhoucke, and A.~Rabinovich, ``Going deeper with convolutions,'' in
  \emph{IEEE Conference on Computer Vision and Pattern Recognition}, June 2015.

\bibitem{he2015delving}
K.~He, X.~Zhang, S.~Ren, and J.~Sun, ``Delving deep into rectifiers: Surpassing
  human-level performance on imagenet classification,'' in \emph{IEEE
  International Conference on Computer Vision}, 2015, pp. 1026--1034.

\bibitem{duchi2011adaptive}
J.~Duchi, E.~Hazan, and Y.~Singer, ``Adaptive subgradient methods for online
  learning and stochastic optimization,'' \emph{Journal of Machine Learning
  Research}, vol.~12, no. Jul, pp. 2121--2159, 2011.

\bibitem{zeiler2012adadelta}
M.~D. Zeiler, ``Adadelta: an adaptive learning rate method,'' \emph{arXiv
  preprint arXiv:1212.5701}, 2012.

\bibitem{kingma2014adam}
D.~Kingma and J.~Ba, ``Adam: A method for stochastic optimization,'' in
  \emph{International Conference for Learning Representations}, 2015.

\bibitem{timofte2014a}
R.~Timofte, V.~De~Smet, and L.~Van~Gool, ``A+: Adjusted anchored neighborhood
  regression for fast super-resolution,'' in \emph{Asian Conference on Computer
  Vision}.\hskip 1em plus 0.5em minus 0.4em\relax Springer International
  Publishing, 2014, pp. 111--126.

\bibitem{kiku2013residual}
D.~Kiku, Y.~Monno, M.~Tanaka, and M.~Okutomi, ``Residual interpolation for
  color image demosaicking,'' in \emph{2013 IEEE International Conference on
  Image Processing}.\hskip 1em plus 0.5em minus 0.4em\relax IEEE, 2013, pp.
  2304--2308.

\bibitem{zoran2011learning}
D.~Zoran and Y.~Weiss, ``From learning models of natural image patches to whole
  image restoration,'' in \emph{IEEE International Conference on Computer
  Vision}, 2011, pp. 479--486.

\bibitem{levin2011natural}
A.~Levin and B.~Nadler, ``Natural image denoising: Optimality and inherent
  bounds,'' in \emph{IEEE Conference on Computer Vision and Pattern
  Recognition}, 2011, pp. 2833--2840.

\bibitem{kim2015accurate}
J.~Kim, J.~K. Lee, and K.~M. Lee, ``Accurate image super-resolution using very
  deep convolutional networks,'' in \emph{IEEE Conference on Computer Vision
  and Pattern Recognition}, 2016, pp. 1646--1654.

\bibitem{yang2010image}
J.~Yang, J.~Wright, T.~S. Huang, and Y.~Ma, ``Image super-resolution via sparse
  representation,'' \emph{IEEE Transactions on Image Processing}, vol.~19,
  no.~11, pp. 2861--2873, 2010.

\bibitem{vedaldi2015matconvnet}
A.~Vedaldi and K.~Lenc, ``Matconvnet: Convolutional neural networks for
  matlab,'' in \emph{Proceedings of the 23rd Annual ACM Conference on
  Multimedia Conference}, 2015, pp. 689--692.

\bibitem{levin2012patch}
A.~Levin, B.~Nadler, F.~Durand, and W.~T. Freeman, ``Patch complexity, finite
  pixel correlations and optimal denoising,'' in \emph{European Conference on
  Computer Vision}, 2012, pp. 73--86.

\bibitem{burger2013learning}
H.~C. Burger, C.~Schuler, and S.~Harmeling, ``Learning how to combine internal
  and external denoising methods,'' in \emph{Pattern Recognition}, 2013, pp.
  121--130.

\bibitem{huang2015single}
J.-B. Huang, A.~Singh, and N.~Ahuja, ``Single image super-resolution from
  transformed self-exemplars,'' in \emph{IEEE Conference on Computer Vision and
  Pattern Recognition}, 2015, pp. 5197--5206.

\bibitem{dong2015compression}
C.~Dong, Y.~Deng, C.~Change~Loy, and X.~Tang, ``Compression artifacts reduction
  by a deep convolutional network,'' in \emph{IEEE International Conference on
  Computer Vision}, 2015, pp. 576--584.

\end{thebibliography}

\end{document}